\documentclass[lettersize,journal]{IEEEtran}
\pdfoutput=1
\usepackage{amsmath,amsfonts}
\usepackage{amssymb}
\usepackage[ruled,linesnumbered]{algorithm2e}
\usepackage{array}
\usepackage[caption=false,font=normalsize,labelfont=sf,textfont=sf]{subfig}
\usepackage{textcomp}
\usepackage{stfloats}
\usepackage{url}
\usepackage{verbatim}
\usepackage{graphicx}
\usepackage{xspace}
\usepackage[most]{tcolorbox}
\usepackage{cite}
\usepackage{hyperref}
\hypersetup{hidelinks,
	colorlinks=true,
	allcolors=black,
	pdfstartview=Fit,
	breaklinks=true}
\usepackage{stfloats}
\usepackage{hyperref}
\hypersetup{hidelinks,
	colorlinks=true,
	allcolors=blue,
	pdfstartview=Fit,
	breaklinks=true}
\usepackage{amsmath}
\usepackage{booktabs}
\usepackage{comment,color}
\usepackage{multirow}

\makeatother
\usepackage{ragged2e}
\usepackage[explicit]{titlesec}
\usepackage[caption=false, font=normalsize, labelfont=sf, textfont=sf]{subfig}
\UseRawInputEncoding

\usepackage{amsthm}
\newtheorem{theorem}{Theorem}

\begin{document}
\title{OMNIS+: Semantic RAN Slicing and Quantization for Dynamic Split Neural Networks}

\author{Langtian~Qin,~\IEEEmembership{Graduate~Student~Member,~IEEE,} 
Avishkar~Dnyaneshwar~Kawadgave,
Ian~Andrew~Harshbarger,
Le\"ila~Nasraoui,~\IEEEmembership{Senior~Member,~IEEE,} Carla~Fabiana~Chiasserini,~\IEEEmembership{Fellow,~IEEE}, and Marco~Levorato,~\IEEEmembership{Senior~Member,~IEEE} 
\IEEEcompsocitemizethanks{\IEEEcompsocthanksitem
 L.~Qin, A.~D. ~Kawadgave, I.~A.~Harshbarger, and M.~Levorato are with the Donald Bren School of Information and Computer Sciences, University of California, Irvine, USA.
 L.~Nasraoui is with the National School of Computer Science (ENSI), University of Manouba, Tunisia.
 C.\,F.~Chiasserini is with Politecnico di Torino, Italy. C.\,F.~Chiasserini is also with CNIT, Parma, Italy, and Chalmers University of Technology, Sweden.
\IEEEcompsocthanksitem
This work was supported in part by the U.S.\ National Science Foundation under grant CCF 2140154.
\IEEEcompsocthanksitem A preliminary version of this work was presented at the IEEE International Conference on Computer Communications (INFOCOM), 2026~\cite{Qin2026}.
}
}

\maketitle

\begin{abstract}
Edge computing enables resource-constrained devices to execute machine learning (ML) applications via task offloading, with radio access network (RAN) slicing playing a crucial role in provisioning joint communication and computing resources. However, existing RAN slicing frameworks rely on static computing models, limiting their ability to exploit dynamic semantic data representations enabled by modern neural architectures. To address this limitation, we propose OMNIS+, a semantic RAN slicing framework for multi-cell edge computing built upon a new generation of dynamic split neural models. OMNIS+ incorporates dynamic neural compression paired with adaptive quantization for task offloading, providing a versatile decision space across communication payload and compute dimensions. Unlike prior work, we explicitly investigate the interplay between neural compression and information quantization in computer vision tasks. To this end, we design a novel ``box'' quantization scheme that markedly enhances resilience to transmission bit errors across diverse compression rates. Considering partial system observability and distinct node objectives, we formulate two interdependent optimization problems: (i)~mobile devices (MDs) maximize inference accuracy under quality-of-service (QoS) and queue-stability constraints by jointly adapting dynamic split deep neural network (DNN) branches and cell association; and (ii)~each edge server (ES) allocates bandwidth, graphics processing unit (GPU) frequency, and modulation and coding scheme (MCS) to maximize the max--min inference accuracy among associated MDs under cell capacity limits. We solve these problems using a multi-agent distributed optimization framework, where MDs act as causal contextual multi-armed bandit (MAB) agents under backlog-aware Lyapunov control, while each ES executes resource allocation via block coordinate descent (BCD). Compared with state-of-the-art baselines, OMNIS+ reduces the QoS violation probability by up to $1.75{\times}$ while achieving a better accuracy--violation trade-off. The evaluation code is available at \url{https://github.com/qlt315/OMNIS/tree/journal}.
\end{abstract}


\section{Introduction}\label{sec:intro}

\begin{figure}[tbp]
	\centering
	\includegraphics[angle=0,scale=0.42]{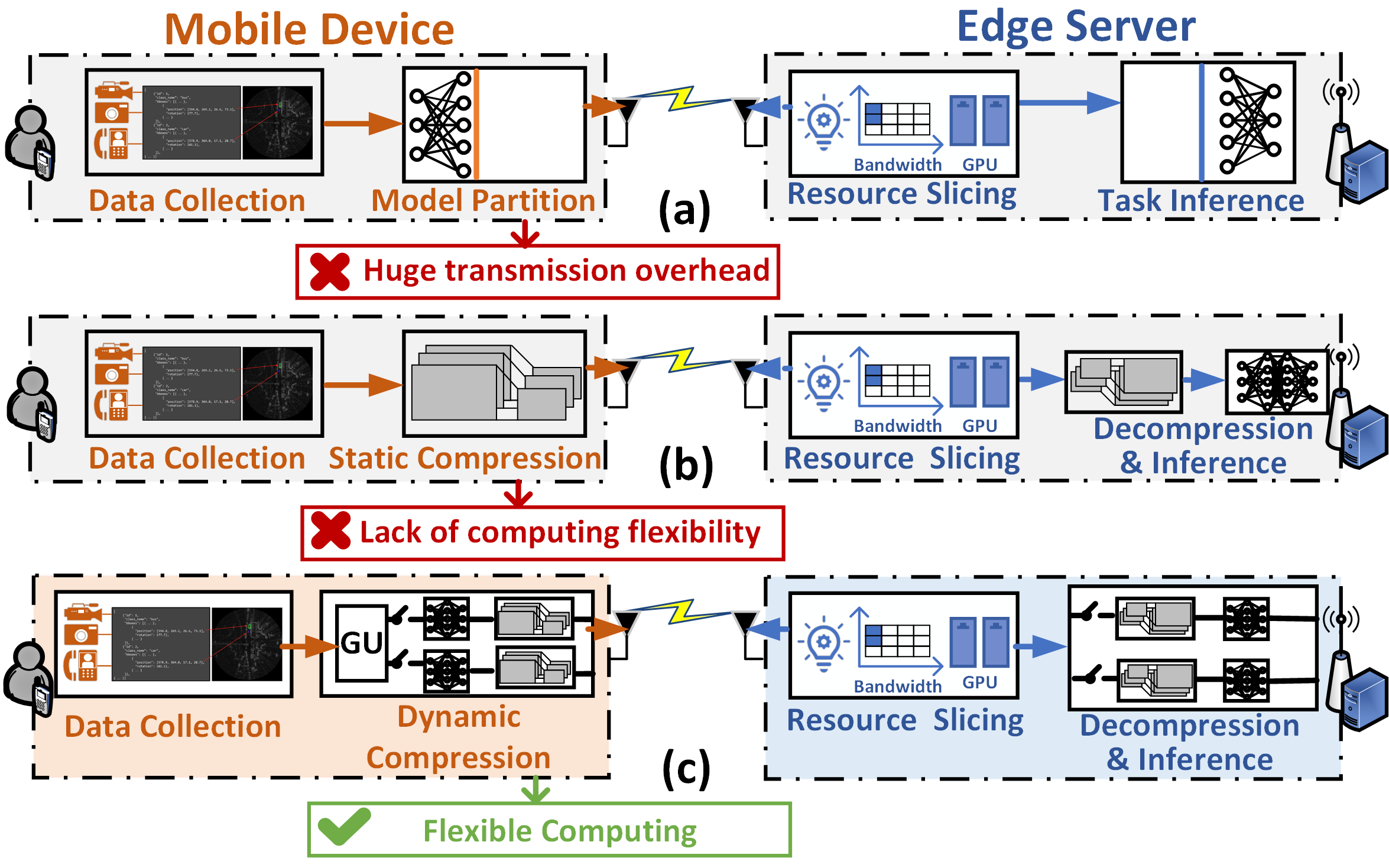}
	\caption{Comparison of existing RAN slicing frameworks for edge computing and OMNIS+: (a) Offloading with neural network partitioning \cite{Xin2021}, (b) Offloading with static data compression \cite{Zelin2024, Corrado2024TMC}, and (c) OMNIS+.}\label{fig:sys_others}
\end{figure}

The rapid proliferation of machine learning (ML)-driven applications--such as real-time video analytics, autonomous driving, and augmented reality--has introduced computationally intensive, latency-critical workloads that involve complex neural models and high-volume sensor data~\cite{Kai2022Mobicom}. Resource-constrained mobile devices (MDs) often struggle to meet these stringent quality-of-service (QoS) demands. To alleviate this bottleneck, mobile edge computing (MEC) offloads computation-heavy ML tasks to edge servers (ESs), significantly reducing execution latency~\cite{Xin2021}. In this context, radio access network (RAN) slicing plays a pivotal role by flexibly provisioning joint communication and computing resources. However, conventional RAN slicing frameworks for edge computing transmit raw sensor data directly to ESs~\cite{Wen2021,Corrado2024ICDCS}, requiring impractically high data rates (e.g., hundreds of kilobits per frame for joint photographic experts group (JPEG) images on the common objects in context (COCO) dataset~\cite{COCO}). Although recent split-computing schemes~\cite{Xin2021} (Fig.\,\ref{fig:sys_others}(a)) partition deep neural networks (DNNs) and transmit intermediate feature tensors from MDs, raw tensors remain bandwidth-prohibitive; for instance, ResNet-50's ``conv3'' feature map requires 3.1\,MB in 32-bit floating-point (FP32) format~\cite{Kaiming2016}.
Alternatively, semantic communication frameworks (Fig.\,\ref{fig:sys_others}(b)) compress data into compact latent representations. Specifically, \cite{Zelin2024} relies on static autoencoders that lack adaptability under fluctuating bit error rates (BERs), whereas \cite{Corrado2024TMC} introduces an adjustable compression factor for dynamic semantic control. Nevertheless, these schemes fail to specify concrete, deployable implementations of flexible compression under realistic wireless channel corruptions.

To address these limitations, we propose OMNIS+, a novel semantic RAN slicing framework tailored for multi-cell edge computing. OMNIS+ connects semantic slicing with dynamic neural networks -- specifically, multi-branch architectures that flexibly map input data into a compact latent space with adaptive quantization. Throughout this paper, \emph{neural compression} refers to the neural-encoder mapping from the input to the latent space, while \emph{compression} denotes the combined neural-compression and quantization process determining the uplink payload size; radio channel coding follows 5G New Radio (5G~NR) modulation and coding scheme (MCS) link adaptation. On the MD side, our dynamic model features branches corresponding to supervised neural encoder--decoder structures chained with multiple quantization schemes  (Fig.\,\ref{fig:sys_others}(c)). These branches offer a fine-grained trade-off between payload compression and error resilience. Upon uplink reception, the ES reconstructs features via the matching neural decoder to perform final inference, forming a flexible, highly efficient semantic execution pipeline when coupled with RAN slicing.

\begin{figure}[tbp]
	\centering
	\includegraphics[angle=0,scale=0.5]{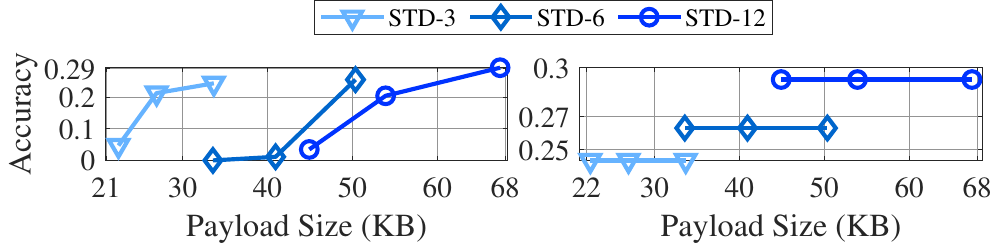}
	\caption{Accuracy vs. payload size under STD quantization at SNR equal to 3\,dB (left) and 10\,dB (right).\label{fig:acc_payload_1}}
\end{figure}

To highlight the impact of feature compression on inference under noisy transmission, we conduct a preliminary study. Fig.~\ref{fig:acc_payload_1} shows instance-segmentation accuracy on the COCO dataset~\cite{COCO} versus payload size for different signal-to-noise ratio (SNR) levels under standard (STD) quantization. Model details and parameters are given in Secs.\,\ref{sec:nn} and\,\ref{sec:eval}. We use a split DNN~\cite{split_computing_survey} with an encoder--decoder for latent neural compression, followed by STD 8-bit integer quantization~\cite{Yoshitomo2022}. The parameters $3, 6$, and $12$ denote the output feature channels of the encoder, where larger values yield lower compression ratios. The compressed tensors are then channel-coded using 5G~NR MCS indices, making the air payload a function of the selected branch $(\beta,\lambda)$ and MCS rate.

This experiment reveals that accuracy scales with payload size up to a saturation plateau, beyond which adding payload bits yields negligible accuracy gains while unnecessarily exacerbating channel overhead--a phenomenon particularly pronounced at high SNR. Conversely, under poor channel conditions, elevated BER corrupts incoming features, severely degrading ES inference accuracy. These insights demonstrate that adaptive semantic compression is essential for building efficient communication--computation pipelines. 

Our main contributions are therefore as follows:

$\bullet$ We introduce OMNIS+, a semantic RAN slicing framework for multi-cell edge computing that leverages a dynamic split DNN architecture to generate flexible, compressed latent data representations optimized for downstream inference.

$\bullet$ We design a novel quantization method that improves the trade-off between compression ratio, bit-error resilience, and task performance in channel regimes where existing quantization techniques fail. To our knowledge, this is the first work to incorporate error resilience into dynamic split DNNs for semantic communications.

$\bullet$ To balance transmission efficiency, inference accuracy, and queue stability, we formulate two interdependent optimization sub-problems for MDs and ESs. MDs maximize a backlog-aware Lyapunov objective by jointly optimizing DNN branch selection and cell association via a causal contextual multi-armed bandit (MAB), while ESs allocate bandwidth, graphics processing unit (GPU) frequency, and MCS via block coordinate descent (BCD) under capacity constraints.

$\bullet$ We perform extensive evaluations using NVIDIA Sionna. Compared with state-of-the-art baselines, OMNIS+ reduces the QoS violation probability by up to $1.75{\times}$ while achieving a better accuracy--violation trade-off.

The rest of the paper is organized as follows. Sec.\,\ref{related_work} reviews related work. Sec.\,\ref{sec:system_model} presents the multi-cell system model, and Sec.\,\ref{sec:nn} details the dynamic split DNN and quantization methods. Sec.\,\ref{sec:methodology} formulates the MD and ES problems and develops the causal contextual MAB and BCD solvers, including regret and complexity analysis in Sec.\,\ref{sec:perf_analysis}. Sec.\,\ref{sec:eval} reports the evaluation, and Sec.\,\ref{sec:conclusions} concludes the paper.

\section{Related work}
\label{related_work}

We review existing work closely related to OMNIS+, categorized into mobile edge computing, RAN slicing, and semantic communications. We also analyze the limitations of prior art and highlight how OMNIS+ overcomes these challenges to achieve highly efficient and adaptive edge intelligence.

{\em Mobile edge computing.}
Recent research focuses on optimizing task offloading from MDs to ESs via either binary or partial offloading paradigms.
In binary offloading, tasks are atomic, requiring full local execution or complete offloading to the edge~\cite{Guangji2023}.
Partial offloading (or split computing) partitions inference tasks, executing initial model layers locally while offloading subsequent layers to the edge~\cite{Liang2023}.
While some studies jointly optimize task offloading and resource slicing~\cite{Langtian2024}, they typically adopt binary offloading models, thereby restricting execution flexibility across communication and computing dimensions.
Conversely, existing partial offloading approaches often assume rigid, non-adaptive task partitioning strategies.
Online control under long-term queue constraints has also been investigated via Lyapunov optimization~\cite{Hongbo2023} and stochastic programming in air--ground edge systems~\cite{Ying2024}; yet, such schemes typically treat model architectures and inference pipelines as static.
OMNIS+ explicitly links dynamic split neural-network selection, neural compression, and quantization directly to offloading decisions, thereby dynamically tailoring the communication--computing trade-off online rather than relying on binary or static partitioning.

{\em RAN slicing.}
It virtualizes physical spectrum and compute resources into isolated network partitions tailored to heterogeneous service demands.
Representative slicing strategies include deep reinforcement learning (DRL) for dynamic caching and mode selection in vehicular networks~\cite{Hongyu2020}; congestion-game formulations with unique Nash equilibrium (NE) properties for distributed, privacy-preserving slicing~\cite{Salvatore2018}; and joint slice association and bandwidth allocation under strict QoS constraints~\cite{Yao2021}.
Other frameworks consider 0-1 quadratic programming for 5G micro-cells~\cite{Salvatore2021}, intent-aware RL for SLA-driven radio scheduling~\cite{Veloso2024}, hierarchical multi-timescale DRL architectures~\cite{Mehdi2022}, and uplink/downlink-decoupled vehicular slicing~\cite{Kai2022}.
However, existing frameworks assume fixed, non-adaptive neural network architectures, failing to exploit key synergies between dynamic model adaptation and resource slicing.
OMNIS+, instead, integrates dynamic neural networks and jointly optimizes multi-cell resource allocation alongside DNN branch control, enabling personalized provisioning and task-specific customization rather than slicing around a static computing pipeline.

{\em Semantic communications.}
They leverage ML-driven data compression to extract and transmit task-relevant information rather than raw bitstreams.
Notably, VQ-DeepSC~\cite{Qifan2023} combines vector quantization with adversarial learning for resilient image transmission; \cite{Danlan2023} applies RL-based adaptive semantic coding to preserve critical task content; and \cite{Jiawei2022} models a rate-distortion trade-off using dynamic feature-length encoding suited for fluctuating channels.
Closest to our work, SEM-O-RAN~\cite{Corrado2024TMC} introduces semantic RAN slicing for edge task offloading, employing a greedy heuristic to allocate radio, compute, and compression resources.
Nevertheless, SEM-O-RAN lacks an explicit feature-compression model and neglects the impact of physical-layer transmission errors and block error rate (BLER) on inference accuracy.
While other semantic designs leave compression unspecified or rely on static architectures, OMNIS+ employs dynamic neural branches with explicit neural compression and quantization while incorporating physical-layer error models into accuracy estimation, significantly enhancing resource utilization over unreliable wireless channels.

\section{System model}\label{sec:system_model}

This section presents the OMNIS+ system model. We first describe the multi-cell RAN and the overall task offloading workflow of OMNIS+. We then detail the local computing, wireless transmission, edge computing, and queueing models.

% \begin{table*}[tbp]
% \centering
% \caption{LIST OF KEY NOTATIONS}
% 	\renewcommand\arraystretch{1.2}
% 	\resizebox{\textwidth}{!}{
% \begin{tabular}{|l|l|}
% \hline
% \textbf{Notation} & \textbf{Definition}  \\ \hline
%  $m,c,t$ & Indices of MD, cell, and time slot \\ \hline
%  $M,C,L,T$ & Numbers of MDs, cells, Top-$L$ candidates, slots \\ \hline
%  $\gamma_{m,c}(t),\hat{\gamma}_{m,c}(t)$ & True / estimated SINR \\ \hline
%  $\beta_m(t),\lambda_m(t),c_m(t),\kappa_m(t)$ & Quantization, latent width, cell, MCS \\ \hline
%  $\alpha_m,\delta$ & Mean arrival rate in tasks/slot, slot duration \\ \hline
%  $d_m(t),b_m(t),B,\eta_m(t),\eta^{u}_m(t)$ & Payload, bandwidth, pool, raw/effective SE \\ \hline
%  $\tau^{l/u/e/s/f/h/b/w}_{m,n}$ & Local/uplink/edge/service/finish/hard/buffer/wait delay \\ \hline
%  $E^{l/u/e/s/h}_{m,n}$ & Local/uplink/edge/service/threshold energy \\ \hline
%  $Q_m(t),A_m(t),S_m(t)$ & Task backlog, arrivals, service \\ \hline
%  $\xi_m(t),\omega^{a},\omega^{q},\beta^{u}$ & accuracy; accuracy/QoS weights; UCB coefficient \\ \hline
%  $u_m(t),V,\Delta_m(t)$ & Utility, Lyapunov weight, drift \\ \hline
%  $\boldsymbol{a},\boldsymbol{K},\mathcal{A},\mathcal{M}_c$ & Vectors/matrices bold; sets calligraphic \\ \hline
% \end{tabular}}
% \end{table*}

\subsection{System overview}
\begin{figure*}[tbp]
	\centering
	\includegraphics[angle=0,scale=0.54]{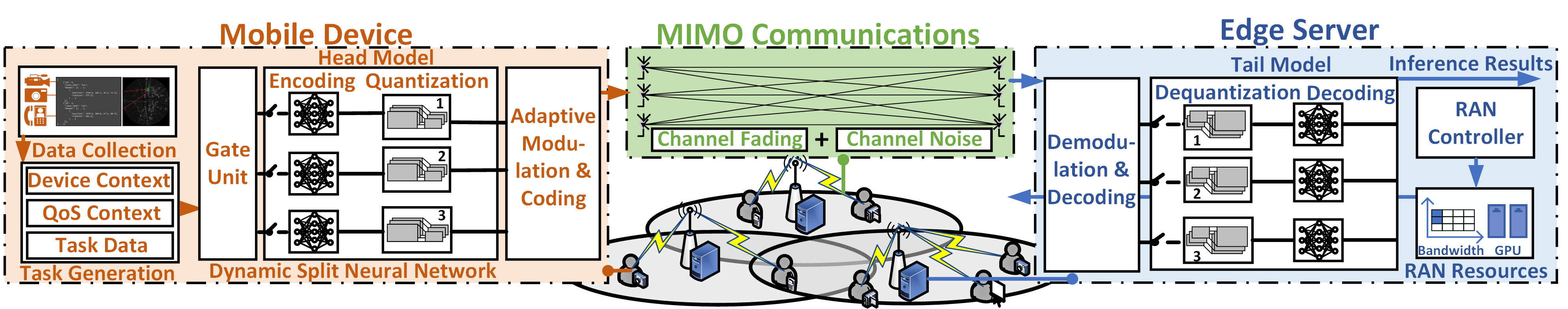}
	\caption{System model of OMNIS+. Randomly arriving tasks are queued at each MD. When a task is admitted, the MD selects a dynamic split-DNN branch and a Top-$L$ serving cell and keeps that association until the task fully completes. Concurrent edge-stage tasks share the cell GPU-frequency pool at the co-located BS/ES. Per-cell RAN control allocates bandwidth, GPU frequency, and MCS.}\label{fig:sys_model}
\end{figure*}
As shown in Fig.\,\ref{fig:sys_model}, OMNIS+ includes an edge computing-enabled multi-cell RAN, consisting of $C$ multi-antenna base stations (BSs) each integrated with an ES and a set $\mathcal{M}=\{1,{\ldots},M\}$ of multi-antenna MDs distributed across the coverage area. Let $\mathcal{C}{=}\{1,{\ldots},C\}$ denote the set of cells. Each MD and each BS are equipped with $\rho^{m}$ and $\rho^{b}$ antennas, respectively. The system operates in time slots indexed by $t=1,{\ldots},T$, each of duration $\delta$. Over this horizon, each MD processes a computer vision job consisting of multiple ML inference tasks. Let $N_m$ denote the number of tasks generated by MD $m$ over $t{=}1,{\ldots},T$, and index these tasks by $n=1,{\ldots},N_m$. Due to limited computing capacity, MD $m$ offloads each task to an ES for processing. Every task is an end-to-end request that may span multiple slots. For instance, in a smart-city video surveillance application, the inference job of each MD is to process a camera stream, with each task corresponding to analyzing a single frame to label people, vehicles, and roadside objects~\cite{Qin2026}. 
Task $n$ of MD $m$ is associated with QoS requirements $\{\tau^{h}_{m,n}, E^{h}_{m,n}, \omega^{d}_{m,n}, \omega^{p}_{m,n}\}$, where $\tau^{h}_{m,n}$ and $E^{h}_{m,n}$ are the target delay and energy budgets. The preference weights $\omega^{d}_{m,n}$ and $\omega^{p}_{m,n}$ capture the relative priority of delay versus energy and satisfy $\omega^{d}_{m,n}{+}\omega^{p}_{m,n}{=}1$. For example, a delay-critical MD such as a drone may raise $\omega^{d}_{m,n}$, whereas an energy-constrained IoT sensor may raise $\omega^{p}_{m,n}$. As tasks and radio conditions vary, a task that misses its delay or energy budget is still completed, and the miss is accounted for in the online utility of Sec.\,\ref{sec:methodology}.

OMNIS+ uses a dynamic split DNN with multiple branches, and a gate unit activates one branch per task. Each branch is a split sub-model with a neural-encoder--decoder design. As detailed in Sec.\,\ref{sec:nn}, different branches trade off compression ratio against resilience to bit errors and inference accuracy. Each branch consists of two parts: the head model, i.e., the neural encoder on the MD side, and the tail model, i.e., the neural decoder on the ES side. The head model that runs on the MD extracts semantic features from the raw input, performs neural compression and quantization, and forms an uplink bitstream via 5G~NR channel coding and modulation. The serving ES recovers the features with the corresponding tail model and produces the inference result, which is then returned to the MD.

In OMNIS+, an optimization engine controls the gate unit and selects a serving cell for each MD from a Top-$L$ candidate set, based on the task context and uplink radio quality. This choice sets the quantization method and the feature-channel width of the active branch. Both the branch and the serving cell are decided once when the task is admitted, and they remain fixed until that task fully completes to avoid frequent cell handover and model switching overhead. At each ES, a RAN controller (i)~allocates bandwidth and selects the MCS for MDs that still have unfinished uplink payloads, and (ii)~shares the GPU-frequency pool among concurrent edge-stage tasks in the cell. Unlike the MD decisions, the RAN slicing strategy is updated every slot as channels, unfinished uplink data, and queues evolve. Thus, as detailed in Sec.\,\ref{sec:methodology}, OMNIS+ couples two-timescale gate control, cell association, and resource slicing in a distributed and context-aware manner.

\subsection{Local computing model\label{sub:MD}}
When MD $m$ admits its $n$-th inference task in slot $t^{a}_{m,n}$, it forms a Top-$L$ candidate set $\mathcal{C}_m(t^{a}_{m,n})$ of nearby cells from the forecast uplink signal-to-interference-plus-noise ratios (SINRs) in dB $\hat{\Gamma}_{m,c}(t^{a}_{m,n})$ (Sec.\,\ref{sec:transmission}). To avoid overcrowded cells, the ranking also uses a coarse load indicator: each ES broadcasts its computing-queue length $Q^{e}_c(t)$, which will be detailed in Sec.\,\ref{sec:queue}. Let $\mathcal{B}$ be the set of quantization methods and $\mathcal{W}$ the set of feature-channel widths. Guided by the optimization engine in Sec.\,\ref{sec:gate_assoc}, the gate unit selects a branch $(\beta_{m,n},\lambda_{m,n})\in\mathcal{B}{\times}\mathcal{W}$ and a serving cell $c_{m,n}\in\mathcal{C}_m(t^{a}_{m,n})$. The raw input is then fed into the head model of the chosen branch, where it undergoes neural compression and quantization by a specific neural network module in Sec.\,\ref{sec:nn}.
The local computing delay of the selected head model is
\begin{equation}
\tau^{l}_{m,n}=\varphi^{l}_{m,n}\,\frac{\psi^{l}_{m,n}}{f^{l}_m \nu^{l}_m\chi^{l}_m},\label{eq:local_delay}
\end{equation}
where $\psi^{l}_{m,n}$ is the floating-point operations (FLOPs) of the head model, and $f^{l}_m$, $\nu^{l}_m$, and $\chi^{l}_m$ are the MD's GPU frequency, number of GPU cores, and FLOPs per cycle per core, respectively. The i.i.d.\ multiplier $\varphi^{l}_{m,n}>0$ is a mean-one lognormal jitter ($\mathbb{E}[\varphi^{l}_{m,n}]=1$) that abstracts mild per-inference hardware variability on the MD including dynamic voltage and frequency scaling (DVFS) micro-adjustments, cache and memory contention, etc. Given the GPU-dependent power coefficient $\zeta^{l}_m$, the local energy consumption is
\begin{equation}
E^{l}_{m,n}=\zeta^{l}_m\bigl(f^{l}_m\bigr)^{3}\tau^{l}_{m,n}
=\varphi^{l}_{m,n}\,\frac{\zeta^{l}_m\bigl(f^{l}_m\bigr)^{2}\psi^{l}_{m,n}}{\nu^{l}_m\chi^{l}_m}.\label{eq:local_energy}
\end{equation}

The resulting uplink bitstream subsequently undergoes channel coding and modulation, as detailed later. The uplink information payload in bits is $d_{m,n}$ and depends on $(\beta_{m,n},\lambda_{m,n})$; the MCS sets the spectral efficiency that drains this payload.

\subsection{Transmission model}\label{sec:transmission}
After local computing, the compressed bitstream is channel-coded and modulated for multiple-input multiple-output (MIMO) uplink transmission. In every slot where MD $m$ still has unfinished uplink data, the serving ES assigns a discrete MCS index $\kappa_m(t)\in\mathcal{K}$ and a bandwidth share $b_m(t)$. In this paper, we adopt the transmission model that aligns with the 5G~NR MIMO uplink standard~\cite{3gpp-ts38214}. Uplink transmission starts at slot $t^{u}_{m,n}$ once the head model finishes all processing. Let $d^{r}_m(t)$ denote the remaining uplink bits in slot $t$, with $d^{r}_m(t^{u}_{m,n})=d_{m,n}$. Within a cell, orthogonal frequency-division multiple access (OFDMA) assigns orthogonal subchannels to associated MDs, so there is no interference from intra-cell MDs, while co-channel interference comes from MDs transmitting to other cells on the same spectrum. Let $p_m$ denote the uplink transmit power of MD $m$, $\mathbf{H}_{m,c}(t)\in\mathbb{C}^{\rho^{b}\times\rho^{m}}$ the MIMO channel from MD $m$ to cell $c$ in slot $t$, $N^{z}$ the noise power spectral density, and $B$ the per-cell bandwidth pool. For the link from MD $m$ to cell $c$, let $\mathcal{I}_{m,c}(t)$ be the set of active interferers, i.e., all MDs $j\neq m$ that upload in slot $t$ to a cell other than $c$. Each interferer $j\in\mathcal{I}_{m,c}(t)$ uses the same fixed-power model with transmit power $p_j$ and channel $\mathbf{H}_{j,c}(t)$ toward cell $c$. Let $\mathcal{M}^{u}_c(t)$ be the set of MDs uploading to cell $c$ in slot $t$. The interference-plus-noise covariance at cell $c$ (absolute power over the allocated band $b_m(t)$) is then
\begin{equation}\label{eq:cov}
\mathbf{R}_{m,c}(t){=}N^{z}b_m(t)\,\mathbf{I}
{+}\hspace{-2mm}\sum_{j\in\mathcal{I}_{m,c}(t)}\hspace{-2mm}\frac{p_j b_m(t)}{B}\,\mathbf{H}_{j,c}(t)\mathbf{H}_{j,c}(t)^{H}.
\end{equation}
For association forecasts before $b_m(t)$ is known, $b_m(t)$ is replaced with an equal share $B/|\mathcal{M}^{u}_c(t)|$. Whitening $\mathbf{R}_{m,c}^{-1/2}\mathbf{H}_{m,c}$ and taking its singular values $\varsigma_{\ell}$ yields $N^{s}=\operatorname{rank}\bigl(\mathbf{R}_{m,c}^{-1/2}\mathbf{H}_{m,c}\bigr)\le\min\{\rho^{b},\rho^{m}\}$ spatial layers. With equal power split across the $N^{s}$ streams, the per-layer SINR of MD $m$ at cell $c$ on layer $\ell$ is
\begin{equation}\label{eq:layer_sinr}
\gamma_{m,c,\ell}(t)=\frac{p_m}{N^{s}}\,\varsigma_{\ell}^{2}(t),\qquad \ell=1,\ldots,N^{s}.
\end{equation}
An exponential effective-SINR mapping (EESM) with parameter $\theta>0$ compresses the layers into a scalar linear effective SINR
\begin{equation}\label{eq:eesm}
\gamma_{m,c}(t)=-\theta\log\Biggl(\frac{1}{N^{s}}\sum_{\ell=1}^{N^{s}}\exp\bigl(-\gamma_{m,c,\ell}(t)/\theta\bigr)\Biggr),
\end{equation}
whose value in dB is $\Gamma_{m,c}(t){=}10\log_{10}\gamma_{m,c}(t)$. Decision-time association uses a forecast $\hat{\gamma}_{m,c}(t)$ of~\eqref{eq:eesm} under the previous slot's transmitter set, with $\hat{\Gamma}_{m,c}(t){=}10\log_{10}\hat{\gamma}_{m,c}(t)$; the environment realizes~\eqref{eq:cov}--\eqref{eq:eesm} with the current radio grants.

Given the serving-cell SINR $\gamma_{m,c_{m,n}}(t)$ and MCS $\kappa_m(t)$, the link produces a spectral efficiency (SE) $\eta_m(t)$  and a BLER $\varepsilon_m(t)\in[0,1]$. The useful spectral efficiency, or goodput, is
\begin{equation}\label{eq:eff_se}
\eta^{u}_m(t)=\bigl(1-\varepsilon_m(t)\bigr)\,\eta_m(t).
\end{equation}
Uploading MDs share the cell bandwidth under $\sum_{m\in\mathcal{M}^{u}_c(t)}b_m(t)\le B$. In each slot the residual payload is drained at rate $b_m(t)\,\eta^{u}_m(t)$:
\begin{equation}\label{eq:uplink_bits}
d^{r}_m(t{+}1)=\bigl[d^{r}_m(t)-b_m(t)\,\eta^{u}_m(t)\,\delta\bigr]^{+}.
\end{equation}
The realized uplink delay $\tau^{u}_{m,n}$ is the accumulated airtime from $t^{u}_{m,n}$ until $d^{r}_m=0$, and the uplink energy is $E^{u}_{m,n}=p_m\tau^{u}_{m,n}$ with the uplink power $p_m$.

\subsection{Edge computing model}\label{sec:edge}
At the ES, the received signal is demodulated and channel-decoded, yielding a recovered bitstream. Edge computing starts only after the uplink tensor has been fully delivered, so that $d^{r}_m=0$. The recovered task then joins the cell computing set defined in Sec.\,\ref{sec:queue}, while the MD keeps its serving-cell choice until the task finishes. Let $\mathcal{M}^{e}_c(t)$ be the set of MDs that have finished uplink to cell $c$ and still have unfinished edge-stage work at that ES in slot $t$. For MD $m\in\mathcal{M}^{e}_c(t)$ with active edge task $n$, let $f^{e}_{m,n}(t)\ge 0$ be the GPU frequency allocated to that task in slot $t$. The ES shares the pool $F^{e}_c$ subject to $\sum_{m\in\mathcal{M}^{e}_c(t)}f^{e}_{m,n}(t)\le F^{e}_c$. Given this allocation, the bitstream is processed by the dequantization module and the neural decoder of the selected tail model to reconstruct features, after which the analysis layers produce the final inference result. Let $\psi^{e}_{m,n}$ be the FLOPs of the matching tail model, and let $\nu^{e}_{c}$, $\chi^{e}_{c}$, and $\zeta^{e}_{c}$ be the number of GPU cores, FLOPs per cycle per core, and power coefficient of the ES in cell $c$. As on the MD, $\varphi^{e}_{m,n}$ is an independent mean-one lognormal factor for mild ES-side runtime jitter. With residual FLOPs $W^{e}_{m,n}(t)$ and $W^{e}_{m,n}(t^{e}_{m,n})=\varphi^{e}_{m,n}\psi^{e}_{m,n}$ at the first edge slot $t^{e}_{m,n}$, the residual drains as
\begin{equation}\label{eq:edge_work}
W^{e}_{m,n}(t{+}1)=\bigl[W^{e}_{m,n}(t)-f^{e}_{m,n}(t)\,\nu^{e}_{c_{m,n}}\chi^{e}_{c_{m,n}}\,\delta\bigr]^{+}.
\end{equation}
The edge delay $\tau^{e}_{m,n}$ is the time from $t^{e}_{m,n}$ until $W^{e}_{m,n}=0$, and the edge energy is
\begin{equation}\label{eq:edge_energy}
E^{e}_{m,n}=\sum_{t:\,W^{e}_{m,n}(t)>0}\zeta^{e}_{c_{m,n}}\bigl(f^{e}_{m,n}(t)\bigr)^{3}\delta.
\end{equation}

The inference result is then returned to MD $m$; we denote by $\xi_{m,n}$ the obtained inference accuracy. Because the downlink payload is much smaller than the uplink feature bitstream, downlink transmission delay is neglected in the total delay.

\subsection{Queueing model}\label{sec:queue}
To capture congestion under random task arrivals and shared edge resources, we maintain a transmission queue at each MD and a computing queue at each ES. All physical queue lengths are counted in the number of tasks. Let $A_m(t)$ be the number of new tasks that arrive and join the transmission queue of  MD $m$ in slot $t$, with mean arrival rate $\alpha_m$. Also, let  $Q^{u}_m(t)$ be  the number of tasks that have arrived at MD $m$ but whose uplink payloads have not yet been fully delivered to a serving ES. This count therefore includes tasks waiting in the buffer and the task under local computing or uplink transmission, if any. Tasks that have already entered the edge stage leave $Q^{u}_m$. When the MD is idle and the transmission buffer is nonempty, it admits the first waiting task, selects a branch and a serving cell, and keeps that choice fixed until the task finishes. On the ES side, multiple edge-stage tasks run concurrently by sharing the GPU-frequency pool  (see Sec.\,\ref{sec:edge}).

After local computing completes, the uplink payload is transmitted to the serving cell. Upon successful delivery, the task leaves $Q^{u}_m$ and enters the computing queue of the serving cell. Let $Q^{e}_c(t)$ denote the number of tasks that are waiting for or receiving edge computing at the ES of cell $c$. When MD $m$ has an unfinished task in slot $t$, let $n_m(t)$ index that task and write $c_m(t)=c_{m,n_m(t)}$ for its locked serving cell. We also define the aggregate congestion indicator
\begin{equation}\label{eq:composite_backlog}
\tilde{Q}_m(t)=Q^{u}_m(t)+Q^{e}_{c_m(t)}(t).
\end{equation}
Thus, the queue updates are
\begin{align}
Q^{u}_m(t{+}1)&=\bigl[Q^{u}_m(t)-C^{u}_{m}(t)\bigr]^{+}+A_m(t),\label{eq:task_queue}\\
Q^{e}_c(t{+}1)&=\bigl[Q^{e}_c(t)-C^{e}_c(t)\bigr]^{+}+A^{e}_c(t),\label{eq:compute_queue}
\end{align}
with $[x]^{+}=\max\{x,0\}$. Here $C^{u}_{m}(t)\in\{0,1\}$ indicates whether MD $m$ finishes uplink delivery in slot $t$, $A^{e}_c(t)=\sum_{m}C^{u}_{m}(t)\mathbb{I}\{c_m(t)=c\}$ aggregates such deliveries into cell $c$, and $C^{e}_c(t)$ is the number of tasks that complete edge computing at cell $c$ in the slot.

\section{Dynamic split DNN model}\label{sec:nn}

%This section presents the dynamic split DNN model at the core of the OMNIS framework, our proposed quantization method, and the trade-offs offered by  OMNIS.

\subsection{Dynamic split DNN architecture}

The architecture of the proposed dynamic split DNN is shown in Fig.\,\ref{fig:sys_model}. It is a gated network: a gate routes each task through one of several sections called ``branches''. The gate selection is controlled by the OMNIS+ optimization engine in Sec.\,\ref{sec:methodology}. Individually, each branch is a split DNN sub-model with $J$ layers and a bottleneck at a fixed split point. This separates the sub-model into a head $\mathcal{H}$ with $l$ layers and a tail $\mathcal{T}$ with $(J-l)$ layers. The head performs neural compression and quantization, mapping the input into a compact bitstream and thereby trading off task accuracy against uplink payload size. We next describe the head, the tail, and the gate unit.

{\em Head model.}
The head model at the MD side includes the set of $l$ initial layers functioning as the neural encoder $\mathcal{H}(x)$, with $x$ being an input sample, for semantic feature extraction. The neural encoder follows the structure in~\cite{Yoshitomo2022} and maps $x$ to an intermediate latent tensor $\boldsymbol{X}$. We do not train $\boldsymbol{X}$ with an explicit latent loss; it is shaped by gradient flow from the reconstruction objective in the tail. A quantization module then follows the encoder to produce a compact bit representation of $\boldsymbol{X}$. As shown in Fig.\,\ref{fig:acc_payload_1} and later in the results, the choice of quantization strongly affects the trade-off among payload size, resilience to bit errors, and inference accuracy.

%Unlike conventional autoencoders, which perform neural compression (followed by reconstruction during neural decoding), the layers in this design are adapted to operate in a trained neural encoder-decoder manner.

%, as 
% \begin{equation}
%    \mathcal{H} =
% \begin{cases}
%     \boldsymbol{h}_0 = \boldsymbol{o}_0 = x, \\
%     \boldsymbol{h}_j = f_j(\boldsymbol{o}_{j-1}, \theta_j), & 1 \leq j \leq l - 1,\\
%     \boldsymbol{h}_{l} = \mathcal{B}(\boldsymbol{o}_{l-1}, \theta_{\mathcal{B}}),
% \end{cases}
% \end{equation}
% where $\theta_{\mathcal{B}}$ is the set of parameters of the bottleneck $\mathcal{B}$ at layer $l$, and $\boldsymbol{o}_{j}$ and $\theta_j$ are the output of the $j$-th layer and the set of parameters of the DNN model, respectively.  

{\em Tail model.}
The tail on the ES inverts the head in two stages. First, an algorithmic dequantizer recovers a tensor representation from the received bitstream. Second, a neural decoder $\mathcal{D}(\cdot)$, trained with teacher--student knowledge distillation~\cite{Yoshitomo2022}, maps the recovered features to a target intermediate representation $h_{j}$ corresponding to the $j$-th layer of a larger, non-split teacher model. The remaining tail layers after this point are copied from the teacher and produce the final inference output.

{\em Gate unit.}
It activates one split-DNN branch per task and thereby determines energy, delay, and accuracy. Concretely, it selects the quantization method $\beta_{m,n}$ and the feature-channel width $\lambda_{m,n}$ from the sets $\mathcal{B}$ and $\mathcal{W}$ defined in Sec.\,\ref{sec:system_model}. In OMNIS+, that choice is scored jointly with Top-$L$ cell association under a single Lyapunov objective, so compression and serving cell are co-optimized. Beyond standard and entropy quantization, we also use our resilient scheme called \textit{box quantization}, detailed next.

\subsection{Box quantization}
\label{Box_quantization}
The quantization module enables dynamic reduction of the transmitted tensor size after the neural encoder. We consider three quantization methods---\textit{standard} (STD), \textit{entropy} (ENT), and the proposed \textit{box}---and detail each below.

\textit{Standard} quantization follows the integer-arithmetic pipeline of~\cite{Jacob2018}, which converts floating-point activations into low-precision integers (INT8) that are efficiently executed on widely available integer-only hardware. Let $\boldsymbol{X}$ denote the tensor output of the neural encoder. A min-max \textit{normalization} maps all floating values in $\boldsymbol{X}$ proportionally into a fixed range so that extreme values do not dominate the representation; the normalized tensor has values between $0$ and $1$. To retain the dynamic range of the original floating-point values, a scaling factor of $(2^{8}-1)$ multiplies the normalized tensor. Finally, a \textit{rounding} operation maps the scaled floating values $\widetilde{\boldsymbol{X}}$ to their nearest discrete integers and yields the quantized tensor $\boldsymbol{X}^{q}=\Big\lfloor\widetilde{\boldsymbol{X}}\Big\rceil$. As we manipulate data at byte level, we fix the bit depth to $8$. Standard quantization achieves only a moderate compression rate, so the resulting payload remains comparatively large and increases uplink overhead.

\textit{Entropy} quantization (e.g., Huffman- or arithmetic-based pipelines, and the rANS backends of~\cite{rANS_entropy,matsubara2022supervised}) can shrink the tensor further by assigning shorter codes to frequent symbols. However, because these codes are variable-length, losing even one bit can drastically change the interpretation of the sequence by shifting the reading frame. This makes entropy quantization unsuitable for distributed edge inference over wireless channels that introduce bit errors. 
To achieve a better trade-off between compression rate and bit-error tolerance, we develop a lightweight and resilient third method that we name \textit{box} quantization. Box quantization first applies the standard pipeline above, and then further compresses $\boldsymbol{X}^{q}$ with an embedding stage that combines run-length embedding (RLE) and index embedding (IE). Prior to embedding, the extremes $\min(\boldsymbol{X})$ and $\max(\boldsymbol{X})$ are prepended as two $32$-bit floating-point values to facilitate de-normalization at the ES. The embedding then proceeds in two stages which are introduced below.

{\em Run-length embedding.}
The quantized tensor $\boldsymbol{X}^{q}$ is first flattened into a vector $\boldsymbol{x}^{q}$ to facilitate sequential processing. RLE then reduces redundancy by storing each run as a (value, frequency) pair in a byte stream, where the first byte is the unique value and the second byte encodes the number of consecutive occurrences. With bit depth $8$, the length of each pair is constrained to $255$, which ensures efficient encoding without exceeding standard byte-size limits. Reshaping $\boldsymbol{x}^{q}$ into value--frequency pairs disrupts the gradient flow in the head model. To help the network learn more compressible representations---and to help the tail decoder correctly recover $\boldsymbol{x}^{q}$---we apply the compression loss
\begin{equation}\label{eq:loss}
    \mathcal{L}^{q}(\boldsymbol{x}^{q}) = \upsilon\,\frac{1}{|\boldsymbol{x}^{q}|}\sum_{i=1}^{|\boldsymbol{x}^{q}|-1}\bigl(x_{i+1}^{q}-x_{i}^{q}\bigr)^{2}\,.
\end{equation}
The parameter $\upsilon$ is a compression scaling factor: larger values favor compressibility at the cost of accuracy, while smaller values prioritize reconstruction fidelity. The loss $\mathcal{L}^{q}(\cdot)$ penalizes squared consecutive differences in $\boldsymbol{x}^{q}$, promoting runs of identical values that shrink the RLE bitstream. Excessive compression, however, can drive $\boldsymbol{x}^{q}$ toward nearly constant values and hurt reconstruction at the tail.

{\em Index embedding.}
IE inserts periodic index markers into the encoded sequence to protect tensor structure under bit errors. Let $\boldsymbol{x}^{c}$ be the RLE output. On the MD, $\iota^{x}$-bit indices are inserted at fixed $\iota^{g}$-bit gaps in the byte stream, mapping the subsequent $8$-bit symbols to positions in $\boldsymbol{x}^{c}$. This adds about $100\iota^{x}/\iota^{g}$ percent overhead. At the ES, the markers allow localized resynchronization after an error, so a corrupted run does not shift the reading frame of the entire payload---a key distinction from entropy quantization. For the air interface, we use 5G~NR MCS coding without upper-layer retransmission of the feature bitstream: residual BLER still reduces goodput in~\eqref{eq:eff_se} and~\eqref{eq:uplink_bits}, while residual bit flips that pass channel decoding enter the decoder and are handled by IE.

{\em Dynamic split DNN training.}
We train the head--tail pair by the two-stage knowledge-distillation procedure of~\cite{Yoshitomo2022}. Stage one freezes the layers after the neural decoder and trains the head plus decoder to match the teacher's $j$-th layer via MSE on $h_{j}$ and intermediate activations; box branches additionally use~\eqref{eq:loss}. Gradients do not pass through RLE/IE or injected bit flips; error resilience at inference relies on IE markers and 5G~NR MCS. Stage two freezes the encoder/decoder and fine-tunes the remaining teacher-initialized tail on the task loss.

\begin{figure}[tbp]
	\centering
	%\captionsetup{singlelinecheck = false, justification=justified}
	\includegraphics[angle=0,scale=0.54]{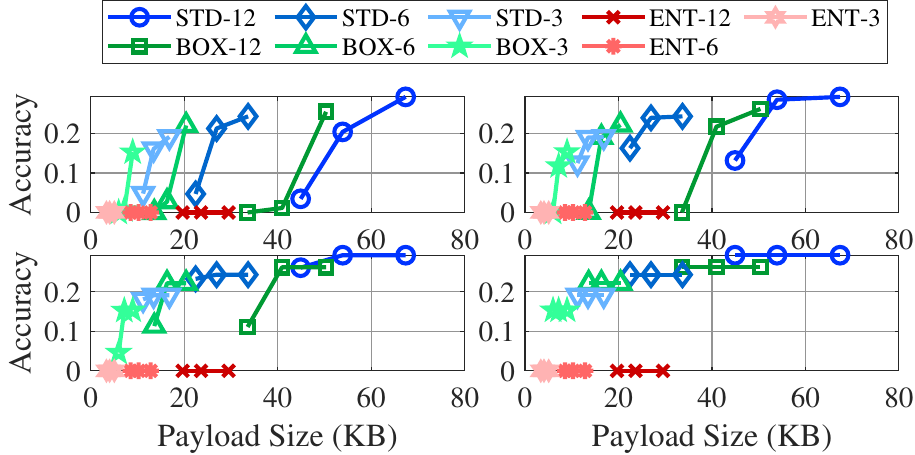}
	\caption{Accuracy vs. payload size for different quantization methods at various SNRs: 3\,dB top-left, 5\,dB top-right, 7\,dB bottom-left, 10\,dB bottom-right.\label{fig:acc_payload_2}}
    
\end{figure}

\subsection{Quantization performance evaluation}
To highlight the impact of different quantization methods on the inference performance, we conduct an evaluation experiment, and depict the accuracy versus payload size under varying SNR values in Fig.\,\ref{fig:acc_payload_2}. In training for our box quantization, we use $\upsilon\in\{0.15, 0.075, 0.0375\}$ for the bottleneck channel sizes of 12, 6, and 3 respectively. We also include entropy quantization (``ENT''), which uses the same neural architecture as the standard branches but replaces the quantizer with a more advanced, partially differentiable entropy backend such as range asymmetric numeral systems (rANS)~\cite{rANS_entropy}. 

Fig.\,\ref{fig:acc_payload_2} provides several key insights. On the one hand, entropy quantization is highly sensitive to bit errors, and even a strong MCS cannot fully protect it. Any presence of bit errors leads to a drastic degradation in accuracy, often driving it close to zero. On the other hand, standard quantization strategies require the largest payload to attain the same accuracy. Box quantization falls in between, achieving a better trade-off between payload size and accuracy. Overall, no quantization scheme guarantees optimal performance under all conditions. This motivates us to formulate an optimization problem to dynamically control the whole pipeline jointly with resource allocation. This is also amplified by multi-cell association and queueing: a branch that is optimal for a lightly loaded strong cell may be infeasible under backlog pressure or a weaker Top-$L$ candidate, so online control must jointly adapt $(\beta,\lambda,c)$ rather than treating quantization as an isolated design choice.

\section{OMNIS+ optimization framework}\label{sec:methodology}
To achieve optimal system performance, gate unit control, cell association, and RAN slicing should be jointly optimized based on task contexts and MD channel conditions. Traditional centralized optimization algorithms adopted by other RAN slicing frameworks suffer from significant drawbacks, including a large solution space which leads to high computational complexity and a high risk of single points of failure. Also, the optimization objectives of MDs and ESs are inherently different. MDs act selfishly, viewing other MDs as competitors for limited resources and focusing on maximizing their own benefits. In contrast, each ES must consider overall fairness within its cell, ensuring a balanced allocation of bandwidth, GPU frequency, and MCS across associated MDs. Due to these factors, formulating a single global optimization problem is impractical. We thus formulate a branch and cell association problem for each MD, and establish a per-cell RAN resource slicing problem for each ES. Decisions follow dual timescales: the quantization method, feature-channel width, and serving cell are chosen only at task admission, while $(b_m(t),\kappa_m(t),f^{e}_{m,n_m(t)}(t))$ are chosen every slot.

\begin{figure}[tbp]
	\centering
	\includegraphics[angle=0,scale=0.11]{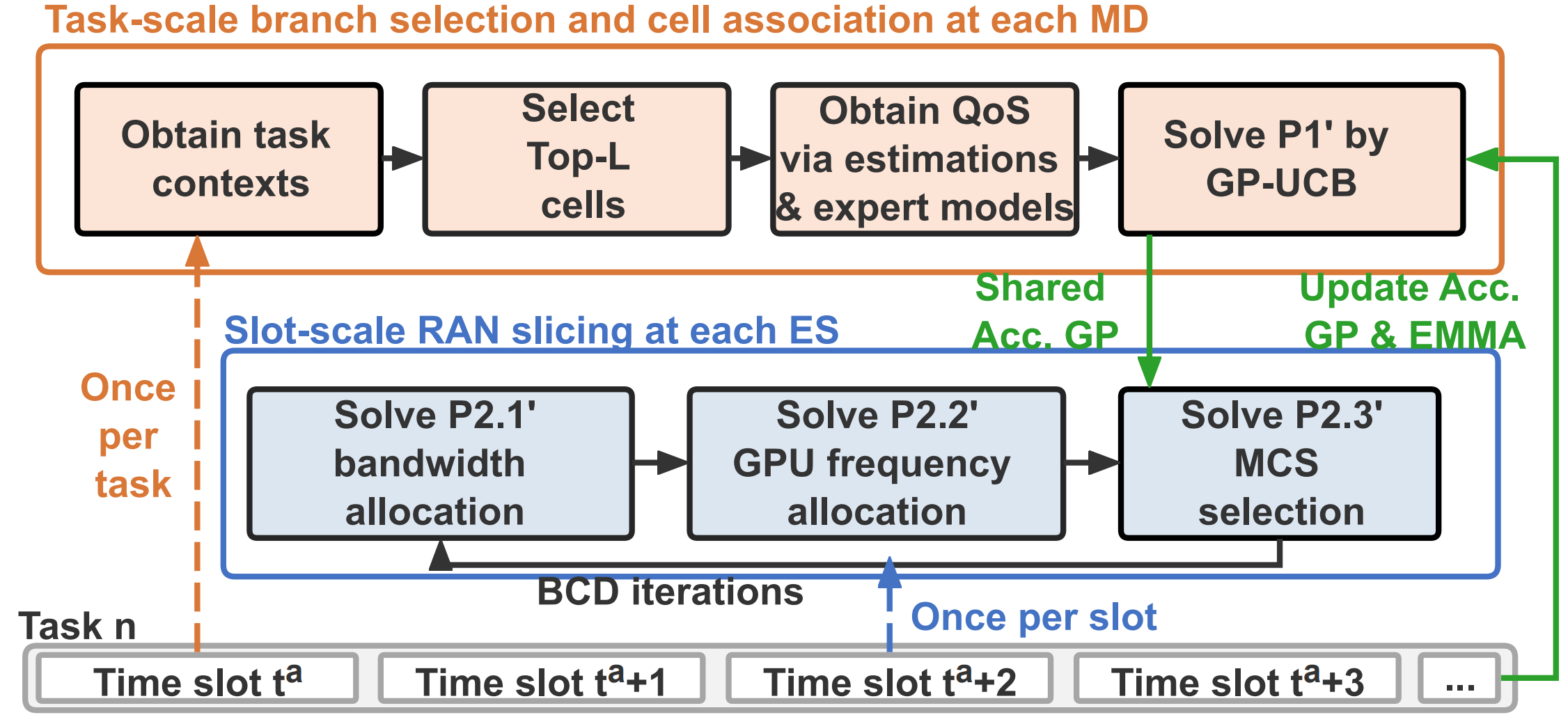}
	\caption{Two-timescale OMNIS+ control flow.}
	\label{fig:algo_framework}
\end{figure}

Fig.\,\ref{fig:algo_framework} provides a roadmap for the OMNIS+ optimization framework. We first cast the MD and ES goals as long-term problems $\mathrm{P1}$ and $\mathrm{P2}$. Because these long-term problems are combinatorial and intractable, we decompose them into instantaneous surrogates $\mathrm{P1'}$ and $\mathrm{P2'}$. At task admission, each MD solves $\mathrm{P1'}$ by Gaussian process upper-confidence-bound (GP-UCB) to determine the branch--cell triple $(\beta,\lambda,c)$. Every slot, each ES solves the per-slot RAN slicing problem $\mathrm{P2'}$ by BCD over three blocks: $\mathrm{P2.1'}$ allocates uplink bandwidth, $\mathrm{P2.2'}$ shares GPU frequency among edge-stage tasks, and $\mathrm{P2.3'}$ selects MCS. A shared accuracy Gaussian process (GP) links MD admission scoring to ES MCS selection, and both sides refresh the GP together with exponentially weighted moving average (EWMA) predictors when a task completes. Below we present these formulations and controllers in detail.

\subsection{Problem formulation: $\mathrm{P1}$ and $\mathrm{P2}$}
We first define the delay, energy, and accuracy metrics used in the optimization problem. For task $n$ of MD $m$, the total end-to-end delay is $\tau^{f}_{m,n}{=}\tau^{b}_{m,n}{+}\tau^{l}_{m,n}{+}\tau^{w}_{m,n}{+}\tau^{u}_{m,n}{+}\tau^{e}_{m,n}$, where $\tau^{b}_{m,n}$ is the queueing delay from arrival till local computing starts, $\tau^{l}_{m,n}$ is the local computing delay, $\tau^{w}_{m,n}$ is the wait from the end of local computing till the next slot boundary, $\tau^{u}_{m,n}$ is the uplink transmission delay, and $\tau^{e}_{m,n}$ is the edge computing delay. Under concurrent GPU sharing, edge execution starts as soon as uplink delivery completes, so ES congestion lengthens $\tau^{e}_{m,n}$ through a smaller share of $F^{e}_c$ rather than adding a separate wait before the tail starts. The total energy is
$E^{f}_{m,n}{=}E^{l}_{m,n}{+}E^{u}_{m,n}{+}E^{e}_{m,n}$, where $E^{l}_{m,n}$ is the MD's local computing energy, $E^{u}_{m,n}$ is the uplink transmit energy, and $E^{e}_{m,n}$ is the edge computing energy on the ES, as given in Secs.\,\ref{sub:MD}--\ref{sec:edge}. The accuracy $\xi_{m,n}$ is obtained when the selected branch is run on the bitstream recovered at the ES upon task completion, under the realized channel trajectory and the MCS that is active when the task finishes.

In the MD problem, each MD aims to maximize the long-term average inference accuracy of completed tasks subject to per-task delay and energy QoS requirements and long-term stability of the MD backlog $\tilde{Q}_m$ in Sec.\,\ref{sec:queue}. Let $K_m(T)$ be the number of tasks completed by MD $m$ by slot $T$. Then
% \begin{tcolorbox}
% [colback=orange!0!white, colframe=orange!70!black, title={Branch and Cell Association Problem on MD $m$},toprule=0mm,titlerule=0mm]
\begin{subequations}\label{P1}
	\begin{align}
		\mathrm{P1}:&\quad
		\max_{\{\beta_{m,n},\lambda_{m,n},c_{m,n}\}}
		\lim_{T\to\infty}\frac{1}{K_m(T)}\sum_{n=1}^{K_m(T)}\xi_{m,n}\\
		\mathrm{s.t.}\;&
		\tau^{f}_{m,n}\le\tau^{h}_{m,n},\quad
		E^{f}_{m,n}\le E^{h}_{m,n},
		\quad\forall n,\label{p1_c1}\\
		&\lim_{T\to\infty}\frac{1}{T}\sum_{t=1}^{T}\mathbb{E}\bigl[\tilde{Q}_m(t)\bigr]<\infty,\label{p1_c2}\\
		&(\beta_{m,n},\lambda_{m,n},c_{m,n})\in\mathcal{B}{\times}\mathcal{W}{\times}\mathcal{C}_m(t^{a}_{m,n}).\label{p1_c3}
	\end{align}
\end{subequations}
% \end{tcolorbox}
\noindent 
Constraint~\eqref{p1_c1} imposes the end-to-end delay and energy budgets of each ML task. Constraint~\eqref{p1_c2} requires strong stability of $\tilde{Q}_m$ (bounded time-average backlog). Constraint~\eqref{p1_c3} restricts each admission decision to a feasible triple.
%: the quantization method $\beta_{m,n}$ must lie in the discrete set $\mathcal{B}$, the feature-channel width $\lambda_{m,n}$ in $\mathcal{W}$, and the serving cell $c_{m,n}$ in the Top-$L$ candidate set $\mathcal{C}_m(t^{a}_{m,n})$ formed at the admission slot $t^{a}_{m,n}$. 
In the online controller of Sec.\,\ref{sec:gate_assoc},~\eqref{p1_c1} is folded into the erf utility of~$\mathrm{P1'}$ so that every admitted task still completes when a budget is missed.

In the RAN slicing problem, given cell associations, each cell allocates bandwidth and MCS to MDs that are still uploading, and shares its GPU frequency pool among MDs whose tasks are already being decoded/inferred at the ES. Each ES maximizes the long-term average of a max--min accuracy score, subject to per-cell capacity. Then,
% \begin{tcolorbox}
% [colback=blue!0!white,colframe=blue!35!black, title={RAN Resource Slicing Problem on ES Cell $c$},toprule=0mm,titlerule=0mm]
\begin{subequations}\label{P2}
	\begin{align}
		\mathrm{P2}:&\quad
		\max_{\boldsymbol{\kappa},\boldsymbol{b},\boldsymbol{f}^{e}}
		\lim_{T\to\infty}\frac{1}{T}\sum_{t=1}^{T}
		\min_{m\in\mathcal{M}_c(t)}\xi_{m,n_m(t)}\\
		\mathrm{s.t.}\;&
		\sum_{m\in\mathcal{M}^{u}_c(t)} b_m(t)\le B,\quad t=1,{\ldots},T,\label{p2_c1}\\
		&
		\sum_{m\in\mathcal{M}^{e}_c(t)} f^{e}_{m,n_m(t)}(t)\le F^{e}_c,\quad t=1,{\ldots},T,\label{p2_c2}\\
		&
		\kappa_m(t)\in\mathcal{K},\quad b_m(t),f^{e}_{m,n_m(t)}(t)\ge 0.\label{p2_c3}
	\end{align}
\end{subequations}
% \end{tcolorbox}
\noindent 
Here $\mathcal{M}_c(t){=}\{m:c_m(t)=c\}$ collects the MDs that currently have an unfinished task at cell $c$, with unfinished-task index $n_m(t)$ as in Sec.\,\ref{sec:queue}. The long-term objective is the time average of the per-slot max--min value, where each active task contributes its eventual completion accuracy $\xi_{m,n_m(t)}$. As in Sec.\,\ref{sec:transmission}, $\mathcal{M}^{u}_c(t)$ is the subset of these MDs that still have residual uplink bits in slot $t$. As in Sec.\,\ref{sec:edge}, $\mathcal{M}^{e}_c(t)$ is the subset whose uplink has already finished and whose edge-stage work is still unfinished in that slot, with GPU frequency $f^{e}_{m,n_m(t)}(t)$ for each such MD. Constraint~\eqref{p2_c1} limits the sum of bandwidth shares of uploading MDs to the cell pool $B$. Constraint~\eqref{p2_c2} limits the sum of GPU frequencies of MDs computing at the ES to the cell GPU pool $F^{e}_c$. Constraint~\eqref{p2_c3} restricts MCS indices to the feasible set $\mathcal{K}$ and keeps bandwidth and GPU allocations nonnegative. It can be seen that $\mathrm{P1}$ is a nonlinear integer program while $\mathrm{P2}$ is a mixed-integer nonlinear program (MINLP). 

\begin{theorem}
\label{nphard}
The per-slot ES problem $\mathrm{P2}$ is NP-hard. Moreover, the one-shot joint branch assignment of all MDs that share a cell under a capacity budget underlying~$\mathrm{P1}$ is NP-hard.
\end{theorem}
\begin{proof}
Both claims follow by reduction from the generalized assignment problem (GAP), which is NP-hard.

For $\mathrm{P2}$, restrict $b_m(t)$ and $f^{e}_{m,n_m(t)}(t)$ to finite grids in a single slot. Together with discrete MCS, each active MD selects one resource bundle. Mapping MDs to jobs, bundles to agents, a sum-of-accuracies surrogate of the max--min score to profits, and~\eqref{p2_c1}--\eqref{p2_c2} to capacities yields a GAP instance, so $\mathrm{P2}$ is NP-hard.

For the joint admission underlying~$\mathrm{P1}$, fix one slot and  cell with association already locked, and let each MD admit one task. Choosing a branch $(\beta,\lambda){\in}\mathcal{B}{\times}\mathcal{W}$ for every MD under a shared bandwidth or GPU budget that meets~\eqref{p1_c1} is again a GAP (jobs=MDs, agents=branches, profit=accuracy, capacity=cell pool). The distributed per-MD controller inherits this coupling by forecasts of the shared pools at admission.
\end{proof}

Beyond combinatorial complexity, several structural challenges render problems $\mathrm{P1}$ and $\mathrm{P2}$ particularly intractable in practice. (1) Inference accuracy $\xi_{m,n}$ depends on the chosen branch, wireless channel trajectory, and final MCS, lacking a closed-form expression under imperfect channel state information (CSI) and non-zero BLER; the ES has no accuracy lookup table, so both MDs and ESs must estimate $\xi$ from a shared GP updated by completion reports. (2) Neither the MD nor the ES possesses exact delay or energy models at decision time, because end-to-end delay and energy closely couple local computing, multi-slot uplink goodput, concurrent GPU sharing, and the hardware architectures and DNN parameters. (3) Each MD observes only local queues, QoS requirements, and Top-$L$ forecast SINRs, forcing it to forecast downstream ES bandwidth, MCS, and GPU allocations that remain unrevealed at admission. Thus, how MDs and ESs share accuracy learning and exchange the information needed to score delay and energy is as vital as the optimization algorithms themselves.

To tackle these issues, we instantiate the two-timescale pipeline of Fig.\,\ref{fig:algo_framework}. First, we reformulate the long-term MD objective, the budgets in~\eqref{p1_c1}, and the queue-stability constraint into a per-admission problem $\mathrm{P1'}$ (Sec.\,\ref{sec:gate_assoc}), solved by a causal contextual MAB that learns accuracy while scoring delay, energy, and backlog online. In parallel, each ES solves a per-slot problem $\mathrm{P2'}$ via BCD over bandwidth, GPU frequency, and MCS (Sec.\,\ref{sec:ran_es}), using a shared accuracy GP together with delay/energy forecasts.

\subsection{Problem transformation: $\mathrm{P1'}$ and $\mathrm{P2'}$}
 $\mathrm{P1}$ maximizes a long-term average of inference accuracy subject to the delay and energy budgets in~\eqref{p1_c1} and to stability of the composite backlog $\tilde{Q}_m$. To obtain a tractable per-admission controller, we fold~\eqref{p1_c1} into a bounded utility and add a backlog-aware score that favors faster service when $\tilde{Q}_m$ is large. 
We encode delay and energy with the error function $\operatorname{erf}$, which is odd, strictly increasing, and bounded. When $\tau^{f}_{m,n}$ or $E^{f}_{m,n}$ lies inside the corresponding budget the error-function term is positive; when it exceeds the budget the term is negative. The utility of task $n$ on MD $m$ is
\begin{equation}\label{eq:utility}
\begin{aligned}
u_{m,n}={}&\omega^{a}\,\xi_{m,n}
+\omega^{q}\Big(
\omega^{d}_{m,n}\,\operatorname{erf}\bigl(\tau^{h}_{m,n}-\tau^{f}_{m,n}\bigr)\\
&+\omega^{p}_{m,n}\,\operatorname{erf}\bigl(E^{h}_{m,n}-E^{f}_{m,n}\bigr)\Big),
\end{aligned}
\end{equation}
where $\omega^{a}$, $\omega^{q}$ weight accuracy vs.\,the QoS score, and $\omega^{d}_{m,n}$, $\omega^{p}_{m,n}$ keep the delay-versus-energy preference of Sec.\,\ref{sec:system_model}. 

The backlog-aware score used at admission is
\begin{equation}\label{eq:backlog_score}
\Delta_m(t)=\tanh\!\bigl(\tilde{Q}_m(t)/Q^{0}\bigr)\bigl(\tilde{U}_m(t)-\tilde{N}_m(t)\bigr),
\end{equation}
where $Q^{0}{>}0$ is a reference backlog counted in tasks, $\tilde{N}_m(t){=}\alpha_m/Q^{0}$ is the mean arrival rate in units of $Q^{0}$, $\tau^{\mathrm{svc}}_{m,n}{=}\tau^{l}_{m,n}{+}\tau^{w}_{m,n}{+}\tau^{u}_{m,n}{+}\tau^{e}_{m,n}$ is the service delay after admission, and $\tilde{U}_m(t){=}\delta/(Q^{0}\tau^{\mathrm{svc}}_{m,n})$ is the corresponding service rate. The factor $\tanh(\tilde{Q}_m(t)/Q^{0})$ is small when the composite queue holds far fewer than $Q^{0}$ tasks and saturates near one once $\tilde{Q}_m$ reaches a few times $Q^{0}$, so $\Delta_m(t)$ stays on a similar numerical scale to $u_{m,n}$. With weights $V{>}0$ and $g^{d}{>}0$, the per-admission problem is
\begin{subequations}\label{P1'}
	\begin{align}
		\mathrm{P1'}:&\quad
		\max_{\{\beta_{m,n},\lambda_{m,n},c_{m,n}\}}
		V u_{m,n}+g^{d}\Delta_m(t)\\
		\mathrm{s.t.}\;&
		(\beta_{m,n},\lambda_{m,n},c_{m,n})\in\mathcal{B}{\times}\mathcal{W}{\times}\mathcal{C}_m(t^{a}_{m,n}).\label{p1p_c3}
	\end{align}
\end{subequations}

For the ES, the per-slot problem $\mathrm{P2'}$ keeps the max--min accuracy structure of $\mathrm{P2}$ and adds a soft delay/energy score built from the sign-flipped erf relative to~\eqref{eq:utility},
\begin{equation}\label{eq:Phi}
\begin{aligned}
\Phi_m(t)={}&
\omega^{d}_{m,n_m(t)}\operatorname{erf}\bigl(\hat{\tau}^{f}_m(t)-\tau^{h}_{m,n_m(t)}\bigr)\\
&+\omega^{p}_{m,n_m(t)}\operatorname{erf}\bigl(\hat{E}^{f}_m(t)-E^{h}_{m,n_m(t)}\bigr),
\end{aligned}
\end{equation}
with the delay and energy forecasts of Sec.\,\ref{sec:ran_es}. Accuracy uses the shared GP of Sec.\,\ref{sec:md_control}: for locked branch $(\beta_{m,n_m(t)},\lambda_{m,n_m(t)})$ and MCS $\kappa$,
$\boldsymbol{z}_m(t,\kappa)=\bigl(\Gamma_{m,c}(t),\beta_{m,n_m(t)},\lambda_{m,n_m(t)},r_{\kappa}\bigr)$ and $\hat{\xi}_m(t,\kappa)=\mu_{N^{\xi}}\bigl(\boldsymbol{z}_m(t,\kappa)\bigr)$ (posterior mean). For $m\in\mathcal{M}^{e}_c(t)$, $\hat{\xi}_m$ is held fixed because the uplink is already complete. The ES solves
\begin{subequations}\label{P2'}
	\begin{align}
		\mathrm{P2'}:&\quad
		\max_{\boldsymbol{\kappa},\boldsymbol{b},\boldsymbol{f}^{e}}
		\min_{m\in\mathcal{M}_c(t)}\hat{\xi}_m(t,\kappa_m(t))
		-\sum_{m\in\mathcal{M}_c(t)}\Phi_m(t)\\
		\mathrm{s.t.}\;&
		\eqref{p2_c1},\eqref{p2_c2},\eqref{p2_c3}.
	\end{align}
\end{subequations}

\subsection{Branch and cell association control on MDs}\label{sec:gate_assoc}\label{sec:md_control}

Once the MD selects action $\boldsymbol{a}_{m,n}=(\beta_{m,n},\lambda_{m,n},i)$, which maps to the $i$-th cell in the Top-$L$ set, its decision remains fixed until the task completes. Since the objective value of $\mathrm{P1'}$ is observed only after task completion, future task admissions can learn from this outcome. Therefore, we formulate $\mathrm{P1'}$ as a contextual MAB problem: at each admission, based on its current observation, i.e., the context, the MD selects an action from $\mathcal{A}=\mathcal{B}{\times}\mathcal{W}{\times}\{1,\ldots,L\}$, with the corresponding reward defined by the Lyapunov score $V u_{m,n}+g^{d}\Delta_m(t)$. Then the MD will know the quality of the selected action and update its policy accordingly.

In OMNIS+, we propose a multi-agent causal contextual MAB framework to solve $\mathrm{P1'}$. Choosing a branch and cell is a causal intervention $do(\beta,\lambda,c)$ given the observed uplink SINR. The MD cannot lock MCS: a trial index $\hat{\kappa}$ is probed from $\hat{\gamma}$ by the same inner-loop link adaptation that the ES uses. Accuracy is learned by a shared GP as a function of $P\bigl(\xi\mid do(\beta,\lambda,c),\Gamma,r\bigr)$, where $r$ is an MCS coding rate; delay and energy are scored by expert models from payload size, predicted goodput, bandwidth, and GPU share. Using the GP only for accuracy keeps its input low-dimensional.

Next we define the context, the arms, and the reward of this causal bandit, and then describe how the policy is learned.

{\em Context}. Before MD $m$ selects a branch and a cell for a task admitted at slot $t=t^{a}_{m,n}$, it constructs a context from all currently available observations: the delay budget $\tau^{h}_{m,n}$, energy budget $E^{h}_{m,n}$, preference weights $\omega^{d}_{m,n}$ and $\omega^{p}_{m,n}$, the composite backlog $\tilde{Q}_m(t)$, the broadcast computing-queue lengths $Q^{e}_c(t)$, forecast uplink SINRs $\hat{\Gamma}_{m,c}(t)$, and elapsed buffer waiting time $\tau^{b}_{m,n}$. The composite backlog follows~\eqref{eq:composite_backlog}, except that at admission the serving cell is not yet locked: the MD uses $Q^{u}_m(t)+Q^{e}_{c}(t)$ for each candidate cell $c$ when that cell is scored. The Top-$L$ set uses two ranking indicators, the forecast SINR $\hat{\Gamma}_{m,c}(t)$ and the broadcast queue $Q^{e}_c(t)$ in tasks. Because these quantities have different units, the MD converts each of them to a standardized score across the nearby cells it can hear (a zero standard deviation is replaced by one) and sums the standardized score of $\hat{\Gamma}_{m,c}(t)$ with that of $-Q^{e}_c(t)$. It then keeps the $L$ cells with the largest sums as $\mathcal{C}_m(t)=\{c_{m,1},\ldots,c_{m,L}\}$ in that order.

Delay and energy use measurement-driven EWMA predictors with smoothing $\phi\in(0,1)$, maintained per branch $(\beta,\lambda)$. On completion of branch $(\beta,\lambda)$, let $\bar{f}^{e}$ be the time-average GPU frequency granted to that task over its edge slots. The MD updates $\hat{\tau}^{l}_{\beta,\lambda}{\leftarrow}(1{-}\phi)\hat{\tau}^{l}_{\beta,\lambda}+\phi\tau^{l}_{m,n}$, $\hat{E}^{l}_{\beta,\lambda}{\leftarrow}(1{-}\phi)\hat{E}^{l}_{\beta,\lambda}+\phi E^{l}_{m,n}$, the edge EWMA coefficients $\hat{\Omega}^{e}_{\beta,\lambda}{\leftarrow}(1{-}\phi)\hat{\Omega}^{e}_{\beta,\lambda}+\phi\tau^{e}\bar{f}^{e}$ and $\hat{\pi}^{e}_{\beta,\lambda}{\leftarrow}(1{-}\phi)\hat{\pi}^{e}_{\beta,\lambda}+\phi E^{e}/(\bar{f}^{e})^{2}$ (distinct from residual FLOPs $W^{e}$ and power coefficients $\zeta^{e}$ in~\eqref{eq:edge_work}--\eqref{eq:edge_energy}), and a provisional bandwidth $\hat{b}_{\beta,\lambda}{\leftarrow}(1{-}\phi)\hat{b}_{\beta,\lambda}+\phi b$ from grants, capped by an equal share of $B$. Under a forecast GPU share $\hat{f}^{e}$, the predictors recover $\hat{\tau}^{e}=\hat{\Omega}^{e}_{\beta,\lambda}/\hat{f}^{e}$ and $\hat{E}^{e}=\hat{\pi}^{e}_{\beta,\lambda}(\hat{f}^{e})^{2}$.

For each candidate $(c,\beta,\lambda)$, a trial MCS $\hat{\kappa}$ is probed by 5G~NR link adaptation under $\hat{\gamma}_{m,c}(t)$~\cite{3gpp-ts38214} (BLER target $\varepsilon^{t}$; highest spectral efficiency among feasible indices). With the table BLER $\hat{\varepsilon}$ and spectral efficiency $\eta_{\hat{\kappa}}$ at $(\hat{\gamma}_{m,c}(t),\hat{\kappa})$, the predicted goodput is $\hat{\eta}^{u}=(1-\hat{\varepsilon})\eta_{\hat{\kappa}}$ from~\eqref{eq:eff_se}. With equal-share GPU $\hat{f}^{e}=F^{e}_c/\max\{1,Q^{e}_c(t)+1\}$, the forecasts are $\hat{\tau}^{u}=d_{m,n}/(\hat{b}_{\beta,\lambda}\,\hat{\eta}^{u})$, $\hat{E}^{u}=p_m\hat{\tau}^{u}$, $\hat{\tau}^{e}=\hat{\Omega}^{e}_{\beta,\lambda}/\hat{f}^{e}$, $\hat{E}^{e}=\hat{\pi}^{e}_{\beta,\lambda}(\hat{f}^{e})^{2}$, then $\hat{\tau}^{\mathrm{svc}}=\hat{\tau}^{l}_{\beta,\lambda}+\hat{\tau}^{w}+\hat{\tau}^{u}+\hat{\tau}^{e}$, $\hat{\tau}^{f}=\tau^{b}_{m,n}+\hat{\tau}^{\mathrm{svc}}$, and $\hat{E}^{f}=\hat{E}^{l}_{\beta,\lambda}+\hat{E}^{u}+\hat{E}^{e}$, where $\hat{\tau}^{w}$ is the residual wait to the next slot boundary. Substituting $\hat{\tau}^{\mathrm{svc}}$ into~\eqref{eq:backlog_score} gives the drift forecast $\hat{\Delta}(\boldsymbol{a})$, and the erf terms of~\eqref{eq:utility} give $q(\boldsymbol{a})$ in~\eqref{eq:causal_score}; accuracy is filled by the GP below.

{\em Arms}. An arm represents a candidate triple $\boldsymbol{a}_{m,n}=(\beta_{m,n},\lambda_{m,n},i)\in\mathcal{A}$, where $\mathcal{A}=\mathcal{B}{\times}\mathcal{W}{\times}\{1,\ldots,L\}$ with cardinality $|\mathcal{A}|=|\mathcal{B}|{\cdot}|\mathcal{W}|{\cdot}L$. The third coordinate $i$ is only the rank in $\mathcal{C}_m(t^{a}_{m,n})$; the serving cell is the mapped identity $c_{m,n}=c_{m,i}$. The causal intervention is therefore $do(\beta,\lambda,c)$ on the locked branch and cell, not on the rank $i$, and the trial MCS is not part of the intervention. Following the two-timescale design in Sec.\,\ref{sec:system_model}, $\boldsymbol{a}_{m,n}$ is fixed upon admission and remains unchanged during task execution.

{\em Rewards}. Upon completion, the MD observes $\xi_{m,n}$, $\tau^{f}_{m,n}$, $E^{f}_{m,n}$, and the component measurements used above to refresh the EWMA predictors and the shared GP. These also form the realized Lyapunov score of $\mathrm{P1'}$. At admission the estimated reward is $V\bigl(\omega^{a}\hat{\xi}(\boldsymbol{a})+\omega^{q}q(\boldsymbol{a})\bigr)+g^{d}\hat{\Delta}(\boldsymbol{a})$, with $q(\boldsymbol{a})$, $\hat{\Delta}(\boldsymbol{a})$ from the expert models and $\hat{\xi}(\boldsymbol{a})$ from the GP.

{\em Policy learning}. The estimated reward in $\mathrm{P1'}$ depends nonlinearly on the branch--cell selection, while nearby feature vectors yield correlated accuracy because nearby $(\beta,\lambda)$ and nearby trial coding rates induce gradual shifts in payload size and BLER. This correlation allows the MD to infer the performance of unobserved arms from observed ones, thereby expediting exploration. Because the MD must make admission decisions prior to ES resource allocation, we employ the GP-UCB algorithm~\cite{Krause2011,Ayala2021INFOCOM} to learn $P\bigl(\xi\mid do(\beta,\lambda,c),\Gamma,r\bigr)$, rather than modeling the full Lyapunov score directly.

Accuracy features have the common form $\boldsymbol{z}=(\Gamma,\beta,\lambda,r)$: uplink SINR in dB, quantization method, feature-channel width, and MCS coding rate. Queue backlogs and QoS budgets are excluded. At admission the MD evaluates $\boldsymbol{z}(\boldsymbol{a})=(\hat{\Gamma},\beta,\lambda,r_{\hat{\kappa}})$ under the candidate-cell forecast and the trial coding rate; the ES MCS block evaluates $\boldsymbol{z}_m(t,\kappa)=(\Gamma_{m,c}(t),\beta,\lambda,r_{\kappa})$ at the current SINR and each candidate rate; each completion appends the admission feature $\boldsymbol{z}(\boldsymbol{a}_{m,n})$ used when the arm was chosen. Let $GP\bigl(\mu(\boldsymbol{z}),k(\boldsymbol{z},\boldsymbol{z}')\bigr)$ be the shared accuracy estimator, initialized with a zero-mean prior of bounded variance. After $N^{\xi}$ completions, let $\boldsymbol{\xi}_{1:N^{\xi}}$ be the realized accuracies as noisy GP samples with variance $(\sigma^{\xi})^{2}$. The posterior mean and covariance kernel are updated as
\begin{equation}\label{eq:mean_update}
\begin{aligned}
\mu_{N^{\xi}}(\boldsymbol{z})={}&\boldsymbol{k}_{N^{\xi}}(\boldsymbol{z})^{\top} \times\bigl(\boldsymbol{K}_{N^{\xi}}+(\sigma^{\xi})^{2}\boldsymbol{I}\bigr)^{-1}\boldsymbol{\xi}_{1:N^{\xi}},
\end{aligned}
\end{equation}
\begin{equation}\label{eq:kernel_update}
\begin{aligned}
k_{N^{\xi}}(\boldsymbol{z},\boldsymbol{z}')={}&k(\boldsymbol{z},\boldsymbol{z}')-\boldsymbol{k}_{N^{\xi}}(\boldsymbol{z})^{\top}\\
&\times\bigl(\boldsymbol{K}_{N^{\xi}}+(\sigma^{\xi})^{2}\boldsymbol{I}\bigr)^{-1}\boldsymbol{k}_{N^{\xi}}(\boldsymbol{z}'),
\end{aligned}
\end{equation}
where $\boldsymbol{k}_{N^{\xi}}(\boldsymbol{z})=\bigl[k(\boldsymbol{z},\boldsymbol{z}_1),\ldots,k(\boldsymbol{z},\boldsymbol{z}_{N^{\xi}})\bigr]^{\top}$ and $\boldsymbol{K}_{N^{\xi}}=\bigl[k(\boldsymbol{z}_i,\boldsymbol{z}_j)\bigr]_{i,j=1}^{N^{\xi}}$. Using this posterior distribution, the MD can evaluate untried arms under the current SINR.

Following~\cite{Krause2011}, we use a stationary, anisotropic kernel. Let $\boldsymbol{y}$ be the length-scale vector and $\boldsymbol{Y}{=}\operatorname{diag}(\boldsymbol{y})$. The Mahalanobis-like distance between  feature vectors is defined as $\sqrt{(\boldsymbol{z}-\boldsymbol{z}')^{\top}\boldsymbol{Y}^{-2}(\boldsymbol{z}-\boldsymbol{z}')}$. We adopt the anisotropic Mat\'ern kernel~\cite{Ayala2021INFOCOM}, with $\sigma^{k}{>}0$ controlling the kernel amplitude:  
\begin{equation}\label{eq:matern}
    \begin{aligned}
k(\boldsymbol{z},\boldsymbol{z}')&=(\sigma^{k})^{2}\Bigl(1+\sqrt{3(\boldsymbol{z}-\boldsymbol{z}')^{\top}\boldsymbol{Y}^{-2}(\boldsymbol{z}-\boldsymbol{z}')}\Bigr)\\
&\quad\times\exp\Bigl(-\sqrt{3(\boldsymbol{z}-\boldsymbol{z}')^{\top}\boldsymbol{Y}^{-2}(\boldsymbol{z}-\boldsymbol{z}')}\Bigr).
    \end{aligned}
\end{equation}

Given the context and action set, MD $m$ uses UCB as the acquisition function. It sets the accuracy forecast to $\hat{\xi}{=}\mu_{N^{\xi}}(\boldsymbol{z})+\vartheta\sigma_{N^{\xi}}(\boldsymbol{z})$, where $\sigma_{N^{\xi}}(\boldsymbol{z})=\sqrt{k_{N^{\xi}}(\boldsymbol{z},\boldsymbol{z})}$ is the posterior standard deviation and $\vartheta>0$ is a confidence multiplier balancing exploration and exploitation. Let $\varpi>0$ be a feasibility margin and $\mathcal{A}^{v}_{m,n}$ the viable subset of arms with $\hat{\tau}^{f}(\boldsymbol{a})\le\varpi\tau^{h}_{m,n}$ and $\hat{E}^{f}(\boldsymbol{a})\le\varpi E^{h}_{m,n}$; if this set is empty, it is replaced by the arm of smallest predicted violation under the erf scores of~$\mathrm{P1'}$. Substituting the UCB forecast together with the context scores $q(\boldsymbol{a})$ and $\hat{\Delta}(\boldsymbol{a})$ into $\mathrm{P1'}$, the MD selects
\begin{equation}\label{eq:causal_score}
\begin{aligned}
\boldsymbol{a}_{m,n}{=}\arg\hspace{-2mm}\max_{\boldsymbol{a}\in\mathcal{A}^{v}_{m,n}}\Big[
&V\big(\omega^{a}\hat{\xi}(\boldsymbol{a})
{+}\omega^{q}q(\boldsymbol{a})\big)
{+}g^{d}\hat{\Delta}(\boldsymbol{a})\Big],
\end{aligned}
\end{equation}
where $q(\boldsymbol{a})=\omega^{d}_{m,n}\operatorname{erf}\bigl(\tau^{h}_{m,n}-\hat{\tau}^{f}(\boldsymbol{a})\bigr)+\omega^{p}_{m,n}\operatorname{erf}\bigl(E^{h}_{m,n}-\hat{E}^{f}(\boldsymbol{a})\bigr)$ is the weighted QoS score in~\eqref{eq:utility}. The hyperparameter $\vartheta$ can be dynamically scheduled using RKHS-norm bounds and cumulative information gain~\cite{Krause2011}. During initial admissions, under-explored feasible branches are prioritized until sufficient observations are collected.

The GP is updated only at completion: the MD reports $\xi_{m,n}$, and the serving ES appends the feature--label pair above. Intermediate slots do not update the GP.

\subsection{RAN resource slicing on the ES}\label{sec:ran_es}
Before MD $m$ starts the uplink, it reports to the serving ES the local quantities that the ES cannot observe: the delay and energy budgets and preference weights $\{\tau^{h}_{m,n},E^{h}_{m,n},\omega^{d}_{m,n},\omega^{p}_{m,n}\}$, the chosen branch and cell $\boldsymbol{a}_{m,n}$, the local delay and energy $(\tau^{l}_{m,n},E^{l}_{m,n})$, the waiting times $(\tau^{b}_{m,n},\tau^{w}_{m,n})$, and the remaining uplink bits $d^{r}_m$. In each later slot $t$, the ES of cell $c$ measures the current SINRs, updates the remaining bits $d^{r}_m(t)$, and tracks the computing-queue length $Q^{e}_c(t)$, the uploading set $\mathcal{M}^{u}_c(t)$, the edge-computing set $\mathcal{M}^{e}_c(t)$, and the composite backlog $\tilde{Q}_m(t)$ in~\eqref{eq:composite_backlog}. With these inputs, the ES solves $\mathrm{P2'}$ in that slot. The local values reported by the MD stay fixed until the task ends. The ES also keeps a copy of the shared accuracy GP from Sec.\,\ref{sec:md_control} and updates it when a completion report $(\boldsymbol{z},\xi)$ arrives.

When the ES tests $(b,\kappa,f^{e})$, it forecasts remaining delay and energy with the same structure as at admission (Sec.\,\ref{sec:md_control}), but using measured SINR, remaining bits $d^{r}_m(t)$, and the EWMA coefficients $(\hat{\Omega}^{e}_{\beta,\lambda},\hat{\pi}^{e}_{\beta,\lambda})$ of the locked branch $(\beta,\lambda)=(\beta_{m,n_m(t)},\lambda_{m,n_m(t)})$ (shared with the MD via completion reports). The remaining uplink terms are $\hat{\tau}^{u}_m=d^{r}_m(t)/(b\,\eta^{u}_m)$ and $\hat{E}^{u}_m=p_m\hat{\tau}^{u}_m$ with $\eta^{u}_m$ from~\eqref{eq:eff_se}. For edge computing, if $m\notin\mathcal{M}^{e}_c(t)$ the ES uses the full-task EWMA values $\hat{\Omega}^{e}_m(t)=\hat{\Omega}^{e}_{\beta,\lambda}$ and $\hat{\pi}^{e}_m(t)=\hat{\pi}^{e}_{\beta,\lambda}$; if $m\in\mathcal{M}^{e}_c(t)$, it sets the residual work coefficient to the unfinished FLOPs mapped into the same units, $\hat{\Omega}^{e}_m(t)=W^{e}_{m,n}(t)/(\nu^{e}_{c}\chi^{e}_{c})$, and scales $\hat{\pi}^{e}_m(t)=\hat{\pi}^{e}_{\beta,\lambda}\,\hat{\Omega}^{e}_m(t)/\hat{\Omega}^{e}_{\beta,\lambda}$. Then $\hat{\tau}^{e}_m=\hat{\Omega}^{e}_m(t)/f^{e}$ and $\hat{E}^{e}_m=\hat{\pi}^{e}_m(t)(f^{e})^{2}$. Adding the already elapsed uplink and edge time/energy since admission together with the MD-reported local terms yields $\hat{\tau}^{f}_m$ and $\hat{E}^{f}_m$ for $\Phi_m$ in~\eqref{eq:Phi}. If the edge stage has not started, $\Phi_m$ uses the equal-share GPU $F^{e}_c/\max\{1,Q^{e}_c(t)+1\}$.

Problem $\mathrm{P2'}$ mixes continuous bandwidth and GPU variables with a discrete MCS, and these three choices affect each other through goodput and the GP accuracy prediction. We therefore apply a BCD-style alternating heuristic in three steps: update bandwidth, update GPU frequency, and update MCS. In each outer round the ES holds two blocks fixed, updates the third, and repeats until $(\boldsymbol{b},\boldsymbol{f}^{e},\boldsymbol{\kappa})$ stop changing or a maximum number of rounds is reached. The allocation from the previous slot is used as the starting point. We do not claim monotonic descent of a single joint objective across the three blocks.

We first update bandwidth while MCS and GPU stay fixed. During this step, we keep the goodput $\eta^{u}_m$ unchanged, since goodput depends on SINR and MCS,  which are both  fixed here. After obtaining the new bandwidths, we refresh SINR and goodput, since a new $b_m$ changes the noise and interference terms. With $\eta^{u}_m$ held fixed, a larger $b_m$ shortens the predicted uplink delay and lowers the predicted uplink energy. We write the backlog-weighted uplink delay and energy of all uploading MDs under the cell bandwidth limit $B$ as
\begin{subequations}\label{P2.1'}
	\begin{align}
		\mathrm{P2.1'}:&
		\min_{\boldsymbol{b}(t)}
		\sum_{m\in\mathcal{M}^{u}_c(t)}\frac{w^{u}_m(t)}{b_m(t)}\\
		\mathrm{s.t.}\;&
		\sum_{m\in\mathcal{M}^{u}_c(t)}b_m(t)\le B,\quad b_m(t)\ge 0.
	\end{align}
\end{subequations}
The uplink bandwidth weight of MD $m$ is
\begin{equation}\label{eq:dprime}
w^{u}_m(t)=\Bigl(1+\frac{\tilde{Q}_m(t)}{Q^{0}}\Bigr)\frac{d^{r}_m(t)\bigl(\omega^{d}_{m,n_m(t)}+p_m\omega^{p}_{m,n_m(t)}\bigr)}{\eta^{u}_m(t)}.
\end{equation}
The factor $(1+\tilde{Q}_m/Q^{0})$ grows when the backlog is high, so busy MDs get higher priority. The remaining bits $d^{r}_m$ give more weight to a larger payload. The mix $(\omega^{d}+p_m\omega^{p})$ reflects whether the MD cares more about delay or energy, and dividing by $\eta^{u}_m$ turns bits into time. From the uplink delay and energy predictions above, we also have
\begin{equation}\label{eq:dprime_meaning}
\frac{w^{u}_m(t)}{b_m(t)}=\Bigl(1+\frac{\tilde{Q}_m(t)}{Q^{0}}\Bigr)\Bigl(\omega^{d}_{m,n_m(t)}\hat{\tau}^{u}_m+\omega^{p}_{m,n_m(t)}\hat{E}^{u}_m\Bigr),
\end{equation}
so $\mathrm{P2.1'}$ minimizes a backlog-weighted sum of the predicted uplink delay and energy. To solve it, we form the Lagrangian with multiplier $\mu^{b}\ge 0$ for the bandwidth limit,
\begin{equation}\label{eq:bw_lag}
\mathcal{J}(\boldsymbol{b},\mu^{b})=\sum_{m\in\mathcal{M}^{u}_c(t)}\frac{w^{u}_m(t)}{b_m(t)}+\mu^{b}\Bigl(\sum_{m\in\mathcal{M}^{u}_c(t)}b_m(t)-B\Bigr).
\end{equation}
Setting $\partial\mathcal{J}/\partial b_m=0$ gives $b_m=\sqrt{w^{u}_m/\mu^{b}}$. Then using $\sum_m b_m=B$ to remove $\mu^{b}$ yields the closed-form shares
\begin{equation}\label{eq:bw_closed}
b^{o}_m(t)=B\frac{\sqrt{w^{u}_m(t)}}{\sum_{i\in\mathcal{M}^{u}_c(t)}\sqrt{w^{u}_i(t)}}.
\end{equation}
So each MD gets a share of $B$ that grows with $\sqrt{w^{u}_m}$. After this step, the ES updates SINR and goodput with~\eqref{eq:cov}--\eqref{eq:eff_se} under the new bandwidths.

We next update GPU frequencies while MCS and bandwidth stay fixed. The ES divides the GPU pool $F^{e}_c$ among the unfinished edge-stage MDs. A higher frequency shortens edge delay, which falls as $1/f^{e}$, but raises energy, which grows as $(f^{e})^{2}$. This trade-off under the pool limit is
\begin{subequations}\label{P2.2'}
	\begin{align}
		\mathrm{P2.2'}:&
		\min_{\boldsymbol{f}^{e}(t)}
		\sum_{m\in\mathcal{M}^{e}_c(t)}\Biggl(
		\frac{\omega^{d}_{m,n_m(t)}\,\hat{\Omega}^{e}_m(t)}{f^{e}_{m,n_m(t)}(t)}\nonumber\\
		&\qquad+\omega^{p}_{m,n_m(t)}\,\hat{\pi}^{e}_m(t)\bigl(f^{e}_{m,n_m(t)}(t)\bigr)^{2}
		\Biggr)\\
		\mathrm{s.t.}\;& \hspace{-3mm}
		\sum_{m\in\mathcal{M}^{e}_c(t)}f^{e}_{m,n_m(t)}(t){\le} F^{e}_c,\quad f^{e}_{m,n_m(t)}(t){\ge} 0.
	\end{align}
\end{subequations}
The first term is the delay cost (ratio of remaining work coefficient to frequency and weighted by $\omega^{d}$);  the second one is the energy cost weighted by $\omega^{p}$. We first ignore the pool limit and minimize each MD's own cost. If $\omega^{p}_{m,n_m(t)}{=}0$, the MD only cares about delay, and we set its closed-form request to $f^{o}_m{=}F^{e}_c$. If $\omega^{p}_{m,n_m(t)}{>}0$, the frequency that best balances delay and energy is
\begin{equation}\label{eq:gpu_stat}
f^{o}_m=\Biggl(\frac{\omega^{d}_{m,n_m(t)}\hat{\Omega}^{e}_m(t)}{2\omega^{p}_{m,n_m(t)}\hat{\pi}^{e}_m(t)}\Biggr)^{1/3}.
\end{equation}
More remaining work or a larger delay weight raises $f^{o}_m$, and a larger energy weight lowers it. Let $\mathcal{M}^{+}_c(t)=\{m\in\mathcal{M}^{e}_c(t):\omega^{p}_{m,n_m(t)}>0\}$ be the edge MDs with a positive energy weight. If every edge MD belongs to this set and the sum of closed-form requests does not exceed $F^{e}_c$, we keep $f^{e}_{m,n_m(t)}(t)=f^{o}_m$. Otherwise we scale the requests to fit the pool,
\begin{equation}\label{eq:gpu_closed}
f^{e}_{m,n_m(t)}(t)=F^{e}_c\frac{f^{o}_m}{\sum_{i\in\mathcal{M}^{e}_c(t)}f^{o}_i}.
\end{equation}

Finally, we update MCS while bandwidth and GPU stay fixed. MDs that have already finished the uplink no longer need MCS for accuracy, so they are skipped in this step. For each uploading MD $m$, we choose $\kappa_m$ while keeping the MCS of all other MDs fixed. Under that choice, the max--min accuracy part of $\mathrm{P2'}$ compares MD $m$'s own accuracy prediction with the lowest predicted accuracy among the other MDs in the cell, $\check{\xi}_m(t)=\min_{j\in\mathcal{M}_c(t)\setminus\{m\}}\hat{\xi}_j\bigl(t,\kappa_j(t)\bigr)$. If $m$ is the only active MD, we set $\check{\xi}_m(t)=+\infty$. For each MCS $\kappa\in\mathcal{K}$, the ES updates $\eta^{u}_m$ with~\eqref{eq:eff_se}, then uses the delay and energy formulas above to get new $\hat{\tau}^{f}_m$, $\hat{E}^{f}_m$, and $\Phi_m$, and reads the GP mean $\hat{\xi}_m(t,\kappa)=\mu_{N^{\xi}}\bigl(\boldsymbol{z}_m(t,\kappa)\bigr)$. It then solves
\begin{subequations}\label{P2.3'}
	\begin{align}
		\mathrm{P2.3'}:& \hspace{-2mm}
		\max_{\kappa_m(t)\in\mathcal{K}} \hspace{-2mm}
		\min\Bigl\{\hat{\xi}_m\bigl(t,\kappa_m(t)\bigr),\,\check{\xi}_m(t)\Bigr\}
		{-}\Phi_m\bigl(t;\kappa_m(t)\bigr).
	\end{align}
\end{subequations}
The first term favors a higher predicted accuracy that does not fall below the other MDs, and $-\Phi_m$ lowers the score when delay or energy becomes worse. If two MCS values give the same score, the one with higher spectral efficiency is kept. The ES uses only the GP mean here; UCB exploration stays at admission on the MD side. After every uploading MD has been assigned an MCS once, the ES starts the next outer BCD round with the updated MCS and goodput, and repeats the three steps until the allocation stabilizes.

\subsection{Performance analysis}\label{sec:perf_analysis}
In this subsection, we analyze three aspects of the proposed OMNIS+ optimization framework: (i)~the learning regret of the multi-agent causal MAB at admission, (ii)~the MD--ES control-plane interaction that keeps the shared accuracy GP and compute predictors synchronized, and (iii)~the overall computing complexity of GP-UCB admission together with per-slot BCD, including ES-side GP mean queries for MCS.

We first examine OMNIS+ learning regret at task admission, showing that it grows sublinearly with the number of completed tasks. As detailed in Sec.\,\ref{sec:md_control}, both the QoS score $q(\boldsymbol{a})$ and the drift term $\hat{\Delta}(\boldsymbol{a})$ are scored by expert models and treated as known at decision time. The unknown term in~\eqref{eq:causal_score} is the expected accuracy $\xi(\boldsymbol{z})$ of the admission feature $\boldsymbol{z}(\boldsymbol{a})=(\hat{\Gamma},\beta,\lambda,r_{\hat{\kappa}})$. The admission-time Lyapunov score is
\begin{equation}\label{eq:true_lyap_score}
\begin{aligned}
S_m(t,\boldsymbol{a})={}&V\Big(\omega^{a}\,\xi\!\bigl(\boldsymbol{z}(\boldsymbol{a})\bigr)
+\omega^{q}q(\boldsymbol{a})\Big)
+g^{d}\hat{\Delta}(\boldsymbol{a}).
\end{aligned}
\end{equation}
Let $K_m$ be the number of completed tasks of MD $m$, each closing one admission decision, and let $t^{a}_{m,s}$ be the admission slot of its $s$-th completed task. With the feasible set $\mathcal{A}^{v}_{m,n}$ of Sec.\,\ref{sec:gate_assoc} written $\mathcal{A}^{v}_{m,s}$ at that admission, and with $\boldsymbol{a}_{m,s}$ the arm chosen then, the instantaneous and cumulative regrets are
\begin{equation}\label{eq:acc_regret_inst}
R_{m,s}=\max_{\boldsymbol{a}\in\mathcal{A}^{v}_{m,s}}S_m(t^{a}_{m,s},\boldsymbol{a})
-S_m\bigl(t^{a}_{m,s},\boldsymbol{a}_{m,s}\bigr),
\end{equation}
$\tilde{R}_m=\sum_{s=1}^{K_m}R_{m,s}$, and $\tilde{R}=\sum_{m\in\mathcal{M}}\tilde{R}_m$.

\begin{theorem}
\label{thm:regret_bound}
Assume a Mat\'ern-$3/2$ accuracy GP with observation noise variance $(\sigma^{\xi})^{2}$ and UCB multiplier $\vartheta_{K^{t}}(\varrho){>}0$ for failure probability $\varrho{\in}(0,1)$. Accuracy observations are pooled across MDs into one shared GP. Let $K^{t}{=}\sum_{m{\in}\mathcal{M}}K_m$ be the total number of completed tasks. Then, for every $K_m{\ge} 1$,
\begin{equation}\label{regret_bound}
P\!\left(\tilde{R}_m\le V \omega^{a}\sqrt{\frac{8K_m\vartheta_{K^{t}}(\varrho)\Psi_{K^{t}}}{\log\!\bigl(1+(\sigma^{\xi})^{-2}\bigr)}}+2\right)\ge 1-\varrho,
\end{equation}
where $\Psi_{K^{t}}$ is the maximum information gain of the shared GP after $K^{t}$ observations. For feature dimension $\dim(\boldsymbol{z})$\,\cite{Krause2011,Ayala2021INFOCOM},
\begin{equation}\label{eq:info_gain}
\Psi_{K^{t}}=\mathcal{O}\!\left((K^{t})^{\frac{\dim(\boldsymbol{z})(\dim(\boldsymbol{z})+1)}{3+\dim(\boldsymbol{z})(\dim(\boldsymbol{z})+1)}}\log K^{t}\right).
\end{equation}
 By a union bound,~\eqref{regret_bound} holds for all $m\in\mathcal{M}$ at once with probability at least $1-M\varrho$, so $\tilde{R}$ is at most the sum of the $M$ right-hand sides.
\end{theorem}

\begin{proof}
Under the shared GP posterior, standard GP-UCB analysis bounds the cumulative accuracy estimation error on $\boldsymbol{z}$ after $K_m$ decisions of MD $m$~\cite{Krause2011,Ayala2021INFOCOM}. Treating the expert scores $q$ and $\hat{\Delta}$ as given at decision time, the Lyapunov-score regret~\eqref{eq:acc_regret_inst} is at most that accuracy regret scaled by $V\omega^{a}$.
\end{proof}

Next, we model the MD--ES control-plane interaction overhead, corresponding to the \emph{Interaction} latency component in Fig.\,\ref{fig:convergence}. Let $\tau^{z}$ denote the control-plane round-trip time and $r^{i}$ denote the control channel transmission rate. OMNIS+ executes $N^{i}=2$ signaling rounds per time slot, comprising a status reporting phase and a resource allocation phase. Denoting the aggregate uplink and downlink control signaling payload sizes by $s^{u}$ and $s^{d}$ bytes, respectively, the interaction latency is expressed as $\tau^{i}=N^{i}\,\tau^{z}+8(s^{u}+s^{d})/r^{i}$.

Under a distributed deployment, each MD uploads its selected decision arm ($s^{a}$ bytes), realized accuracy feedback ($s^{r}$ bytes) used to append $(\boldsymbol{z},\xi)$ to the shared GP at the serving ES, and a local status report ($s^{l}$ bytes). The serving ES returns a resource grant allocation ($s^{g}$ bytes) and an edge completion notification ($s^{e}$ bytes); when the GP replica is not co-located with every MD, the ES may also push a compact GP-update digest ($s^{y}$ bytes) so that MDs refresh their local UCB posterior. Each cell broadcasts a compute-state message ($s^{c}$ bytes) to facilitate the next Top-$L$ ranking. Thus, the aggregate uplink and downlink signaling sizes are $s^{u}=M(s^{a}+s^{r}+s^{l})$ and $s^{d}=M(s^{g}+s^{e}+s^{y})+C s^{c}$, respectively. Under a centralized deployment of OMNIS+, each MD instead uploads $N^{x}$ raw floating-point observation metrics ($s^{x}$ bytes each), and the central controller---which hosts the single shared GP---returns both the arm decision and resource grant, yielding $s^{u}=M(N^{x}s^{x}+s^{l})$ and $s^{d}=M(s^{a}+s^{g}+s^{e})+C s^{c}$. Consequently, distributed OMNIS+ exchanges compact arm reports and accuracy feedback to keep ES/MD GP replicas aligned, whereas the centralized counterpart uploads full observation vectors and hosts one GP at the controller.

Finally, we analyze the  complexity of the  OMNIS+ optimization framework, which coordinates GP-UCB admission at MDs with per-slot BCD at ESs, where the MCS block scores each grant candidate by a shared-GP posterior mean query.

\begin{theorem}
\label{complexity}
Assume that the per-cell BCD procedure requires at most $I^{b}$ outer iterations per slot. Let $K^{t}=\sum_{m\in\mathcal{M}}K_m$ be the total number of completed tasks and $N^{\xi}\le K^{t}$ the shared accuracy-GP size. Over $T$ time slots, the overall computational complexity of OMNIS+ is
$\mathcal{O}\!\big((N^{\xi})^{3}+K^{t}|\mathcal{A}|N^{\xi}+CT I^{b}\bigl(\bar{M}^{u}(1+|\mathcal{K}|N^{\xi})+\bar{M}^{e}\bigr)\big)$,
where $|\mathcal{A}|=|\mathcal{B}|{\cdot}|\mathcal{W}|{\cdot}L$, and $\bar{M}^{u}$ and $\bar{M}^{e}$ are the maximum numbers of radio-active and concurrent edge-stage MDs in a cell.
\end{theorem}

\begin{proof}
The shared GP is updated once per completion. An incremental posterior update after the sample $n$ costs $\mathcal{O}(n^{2})$, so $N^{\xi}$ completions cost $\sum_{n=1}^{N^{\xi}}\mathcal{O}(n^{2})=\mathcal{O}((N^{\xi})^{3})$. With a cached factorization of $\boldsymbol{K}_{N^{\xi}}+(\sigma^{\xi})^{2}\boldsymbol{I}$, each mean query is $\mathcal{O}(N^{\xi})$; a variance query is $\mathcal{O}((N^{\xi})^{2})$ in general, and we count the dominant mean-query cost for admission scoring, giving $\mathcal{O}(K^{t}|\mathcal{A}|N^{\xi})$. Per cell and slot,~\eqref{P2.2'} costs $\mathcal{O}(\bar{M}^{e})$. Within each BCD outer iteration,~\eqref{P2.1'} costs $\mathcal{O}(\bar{M}^{u})$ and~\eqref{P2.3'} evaluates $|\mathcal{K}|$ candidates per radio-active MD with one GP mean and one $\Phi$ forecast, for $\mathcal{O}(\bar{M}^{u}|\mathcal{K}|N^{\xi})$. Over $I^{b}$ iterates, $C$ cells, and $T$ slots, we get the last term.
\end{proof}
Overall, the complexity is polynomial in the shared-GP size, the admission volume, and the per-slot radio/edge workloads.

\section{Performance evaluation}\label{sec:eval}
\subsection{Parameter settings}
We use NVIDIA Sionna~\cite{Hoydis2022Sionna} to simulate the multi-cell RAN. The wireless channel follows the 3GPP TR~38.901 Urban Micro model at $3.5$\,GHz, and includes large-scale path loss and shadow fading. The simulated deployment has $C{=}7$ omnidirectional cells at the centers of a hexagonal grid with inter-site distance $200$\,m. Each BS uses a $2{\times}4$ array with $8$ receive antennas and an omnidirectional element pattern, while each MD has $2$ transmit antennas. We adopt the random walk model to characterize the mobility, and set the moving speed as $0.8$\,m/s~\cite{Qin2024}. The slot duration $\delta$ is 1\,s. Thermal noise density is $-174$\,dBm/Hz with an $8$\,dB noise figure, over the MD's allocated bandwidth. MCS indices take values in $\{0,{\ldots},28\}$. The BLER target is $0.1$. Each cell offers a bandwidth pool of $B{=}650$\,kHz. Every MD selects a serving cell from a Top-$L{=}3$ candidate set ranked by forecast uplink SINR in dB and the broadcast computing-queue length $Q^{e}_c$.

Each MD uses GPU frequency $1.6$\,GHz, $768$ cores, $12$ FLOPs per cycle, power coefficient $0.35$, and transmit power $0.1$\,W. Each ES has GPU pool $F^{e}_c{=}4.0$\,GHz, $12288$ cores, $24$ FLOPs per cycle, and power coefficient $0.75$. Realized local/edge FLOPs are further scaled by independent mean-one lognormal jitter with scale $0.1$ (about $10\%$ CV), abstracting mild DVFS/runtime variability as in~\eqref{eq:local_delay}--\eqref{eq:edge_energy}. Per-task delay and energy thresholds are drawn from $\mathcal{U}(1.60,2.50)$\,s and $\mathcal{U}(0.11,0.26)$\,J. The energy preference weight satisfies $\omega^{p}_{m,n}{\sim}\mathcal{U}(0.25,0.45)$, and $\omega^{d}_{m,n}=1-\omega^{p}_{m,n}$. Task arrivals are Poisson with mean rate $\alpha_m{=}0.07$ tasks per slot. We set the Lyapunov weight to $V{=}10$, the accuracy weight to $\omega^{a}{=}24$, and the QoS coefficient to $\omega^{q}{=}20$. The drift term is scaled by $g^{d}{=}1.5$. OMNIS+ and C-OMNIS+ use the admission feasibility margin $\varpi{=}0.88$ on predicted delay and energy; GDO defaults to $\varpi{=}1.0$ (the Acc--vio Pareto in Fig.\,\ref{fig:acc_vio} sweeps $\varpi$ for both schemes). The drift terms are normalized by the task-count scale $Q^{0}{=}3$. Per-cell BCD stops at tolerance $10^{-4}$ or after at most $30$ iterations.

The learning task is instance segmentation on the COCO dataset~\cite{COCO} with $80$ categories, reported by mAP@[0.5:0.95]. The dynamic split DNN is built on Mask R-CNN with a ResNet-50 backbone. The neural encoder has three convolutional stages: the first follows the initial ResNet-50 block with batch normalization, a rectified linear unit (ReLU), and max-pooling; the next two stages use stride-$2$ convolutions for downsampling. The tail aligns $h_{j}$ with the output of ResNet-50's first block under the stage-one mean squared error (MSE) reconstruction of Sec.\,\ref{sec:nn}, while the remaining three ResNet blocks stay in the teacher-initialized tail; box branches additionally use~\eqref{eq:loss} during that stage. The decoder has two deconvolutional and two convolutional layers, and box quantization uses a standard bottleneck. Online control selects among the $6$ branches $\{\mathrm{Box},\mathrm{STD}\}{\times}\{3,6,12\}$ and a Top-$L{=}3$ cell, so $|\mathcal{A}|=18$ arms. Entropy quantization is not used online because of its extreme BER sensitivity. Finally, we distill from a Mask R-CNN teacher (mAP 41.80\%) and train the head/tail models using the two-stage procedure in Sec.\,\ref{sec:nn} and~\cite{Yoshitomo2022}.

% \begin{table}[t]
% \caption{Online branch payloads and clean accuracy values used in the simulator.}
% \label{tab:base_model_metrics}
% \centering
% \begin{tabular}{c c c c}
% \hline
% Channels & Quantization & Payload (KB) & Clean accuracy \\
% \hline
% 12 & STD & 44.93 & 0.2928 \\
% 6 & STD & 22.46 & 0.2435 \\
% 3 & STD & 11.23 & 0.1916 \\
% \hline
% 12 & BOX & 33.58 & 0.2623 \\
% 6 & BOX & 13.26 & 0.2217 \\
% 3 & BOX & 6.00 & 0.1532 \\
% \hline
% \end{tabular}
% \end{table}

\subsection{Benchmarks}
We compare OMNIS+ against the following benchmarks, modified or extended from state-of-the-art RAN slicing or edge computing frameworks. We label the proposed scheme as \textbf{``OMNIS+''} and retain the ablation \textbf{``OMNIS''} from our previous work~\cite{Qin2026}. OMNIS fits a contextual GP to the scalar utility $\hat{u}_m$ and selects arms by $V\bigl(\mu(\boldsymbol{a})+\vartheta\sigma(\boldsymbol{a})\bigr)+g^{d}\Delta$ with UCB acquisition, where $\mu$ and $\sigma$ are the posterior mean and standard deviation of that utility GP. Unlike OMNIS+, which estimates accuracy through a causal mechanism $P\bigl(\xi\mid do(\beta,\lambda,c),\Gamma,r_{\hat{\kappa}}\bigr)$ and evaluates soft QoS with the wireless model and measurement-driven compute predictors before composing with Lyapunov drift, OMNIS learns $\hat{u}_m$ end-to-end, so accuracy and QoS effects remain entangled in the GP while drift stays an additive online term. Other benchmarks are introduced as follows.

$\bullet$ {\em Greedy-based optimization (GDO)}~\cite{Corrado2024TMC, Xin2021}: GDO employs a heuristic for joint branch and Top-$L$ cell selection. For each candidate arm it predicts delay and energy with the \emph{same} measurement-driven compute predictors and wireless model as OMNIS+. An arm is QoS feasible if those predictions meet the task delay and energy limits. Accuracy is estimated separately. For each branch, GDO maintains a running average of realized accuracy from completed tasks, and adds a small exploration bonus when that branch has been tried few times. Among QoS feasible arms it selects the one with the largest accuracy estimate. Thus GDO differs from OMNIS+ not only in the objective, but also in how accuracy is learned. The ES then solves the same BCD problem as OMNIS+.

$\bullet$ {\em Centralized OMNIS+ (C-OMNIS+)~\cite{Ayala2021INFOCOM, Qin2026}}: C-OMNIS+ is the centralized counterpart of OMNIS+. A central controller hosts the shared accuracy GP and scores each MD with the same causal objective, measurement-driven QoS predictors, drift, and grant-EWMA bandwidth forecast. Scoring and GP updates run sequentially, and MDs upload observations to the controller instead of deciding locally. Each ES still solves the same BCD problem as OMNIS+.

$\bullet$ {\em Deep Q-network (DQN)}~\cite{Mhatre2025, Chen2024}: A centralized value-based DRL baseline that approximates joint action values with a neural network. One learner observes the concatenated global state of all MDs and selects a joint action $a{=}(a_1,{\ldots},a_M)$, where each local action encodes a (branch, Top-$L$ cell-rank) pair. The reward is the mean Lyapunov objective $Vu+g^{d}\Delta$. When training the model, an online network selects the greedy next action, while a periodically synced target network evaluates that action to reduce overestimation, together with experience replay and epsilon-greedy exploration. The ES solves the same BCD problem as OMNIS+.

$\bullet$ {\em Proximal policy optimization (PPO)~\cite{Nguyen2024, Xie2025}}: A centralized DRL baseline that uses an actor--critic structure to learn the arm selection strategy. A shared-trunk actor factors as $\prod_i\pi_i(a_i{\mid}s)$, and a critic estimates the return $Vu+g^{d}\Delta$. Training follows on-policy clipped PPO with generalized advantage estimation (GAE) and entropy regularization. The ES solves the same BCD problem as OMNIS+.

$\bullet$ {\em Multi-agent PPO (MAPPO)~\cite{Qin2024, Li2025}}: MAPPO extends PPO under centralized training with decentralized execution (CTDE). Each MD runs a parameter-shared local actor on its own observation and selects a local action that encodes a branch and a Top-$L$ cell rank. A centralized critic is trained on the concatenated global state of all MDs, with the team reward equal to the mean Lyapunov objective $Vu+g^{d}\Delta$. The ES solves the same BCD problem as OMNIS+.

\subsection{Results and analysis}
We evaluate all schemes under different parameter settings in terms of the Lyapunov reward, delay, energy consumption, inference accuracy, task backlog, and QoS constraint violation probability.

\begin{figure}[tbp]
	\centering
	\includegraphics[width=\columnwidth]{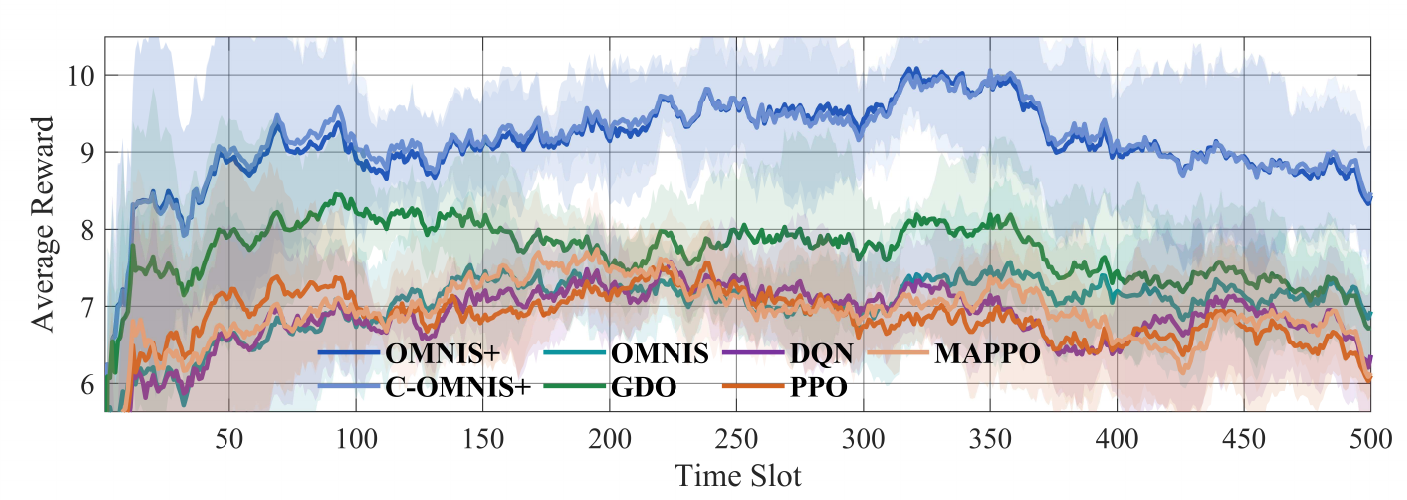}\\
	\includegraphics[width=\columnwidth]{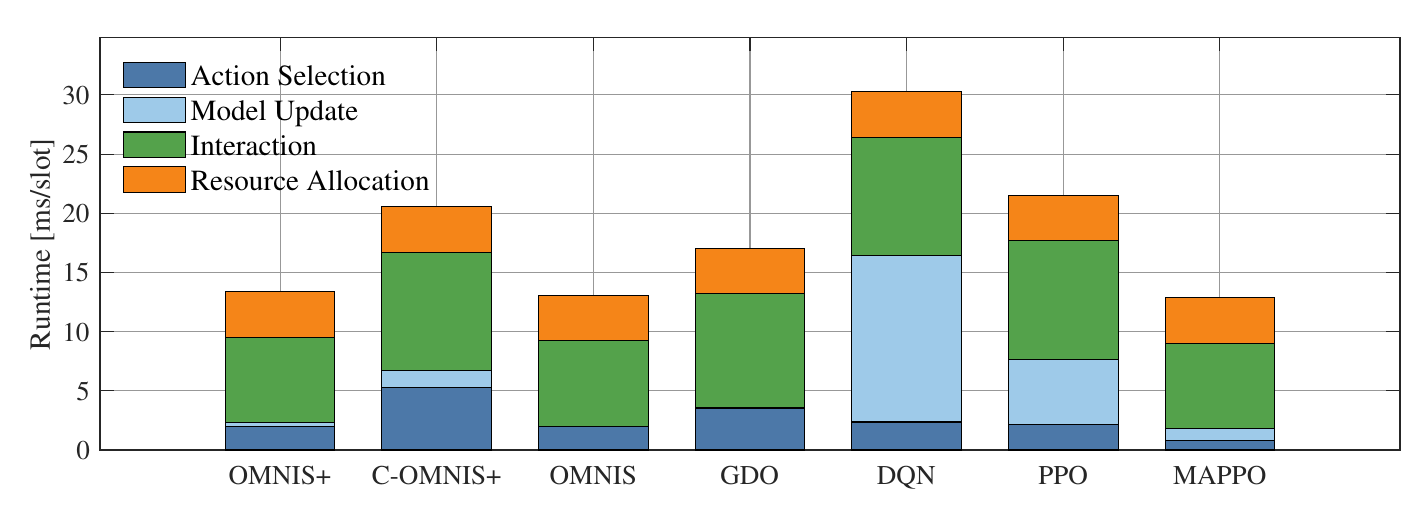}
	\caption{Convergence and running time performance of different schemes. The top panel shows the Lyapunov reward per slot averaged over MDs over $500$ slots. The bottom panel shows control plane running time.}
	\label{fig:convergence}
\end{figure}

{\em Convergence performance.} The top panel of Fig.\,\ref{fig:convergence} reports the Lyapunov reward under the default setting with $M{=}25$, averaged over $5$ random seeds. The plotted horizon is $T{=}500$ slots. Curves are running averages over a trailing window of $60$ slots of the per-slot MD-mean objective, and the shaded bands are mean $\pm$ one standard deviation across seeds. OMNIS+ rises quickly and stays near a late-horizon mean of about $8.4$, the best among distributed schemes. It learns only accuracy with a Gaussian process, and evaluates delay, energy, and queue backlog with the shared measurement-driven predictors, so each MD can raise $Vu+g^{d}\Delta$ once accuracy estimates become reliable. OMNIS stays near $6.9$ because it learns one scalar utility and mixes accuracy with QoS penalties. Relative to OMNIS, OMNIS+ improves this per-slot reward by about $22\%$. C-OMNIS+ follows OMNIS+ closely and ends near $8.5$, about $1\%$ above OMNIS+. The small gap arises because C-OMNIS+ rescores MDs that pick the same cell under a shared bandwidth forecast, while OMNIS+ lets each MD decide locally in parallel. GDO stays near $6.7$. Among arms that meet the delay and energy limits, it simply picks the one with the highest predicted accuracy and never accounts for queue backlog in its objective. DQN, PPO, and MAPPO remain in the $6.1$--$6.4$ range, because a scalar return is hard to learn when accuracy feedback is delayed and the channel across cells keeps changing.

\begin{figure*}[tbp]
	\centering
	\includegraphics[width=\textwidth]{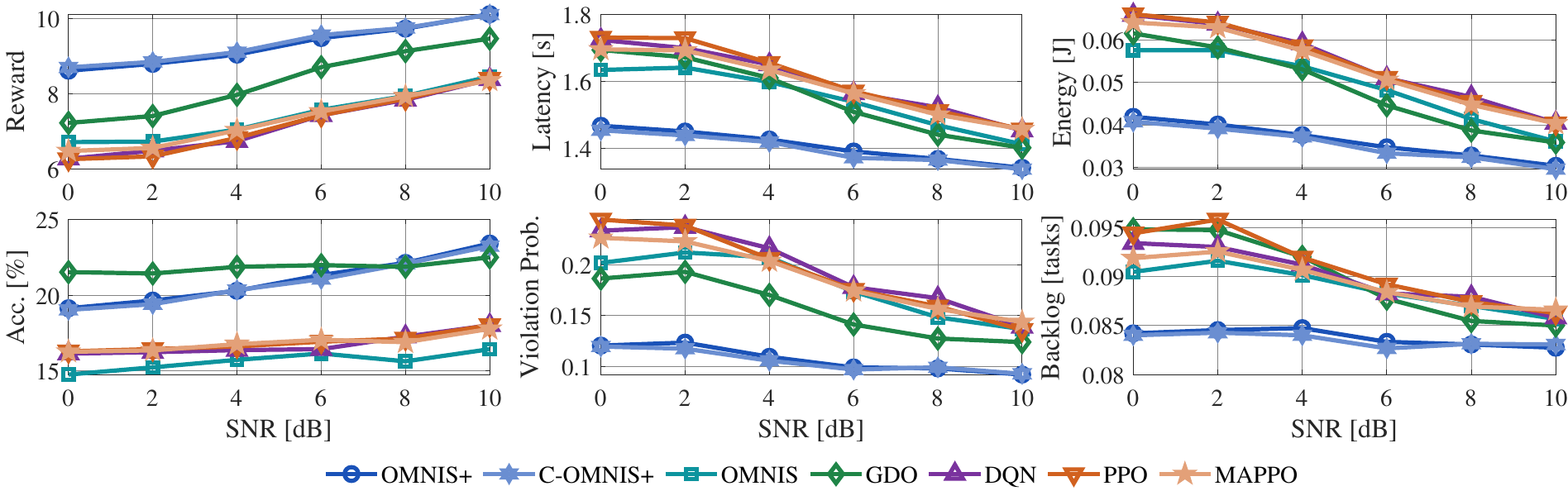}
	\caption{Performance of different schemes under varying SNR. The top row shows reward, average delay, and energy from left to right. The bottom row shows accuracy, violation probability, and backlog from left to right.}
	\label{fig:diff_snr}
\end{figure*}

\begin{figure*}[tbp]
	\centering
	\includegraphics[width=\textwidth]{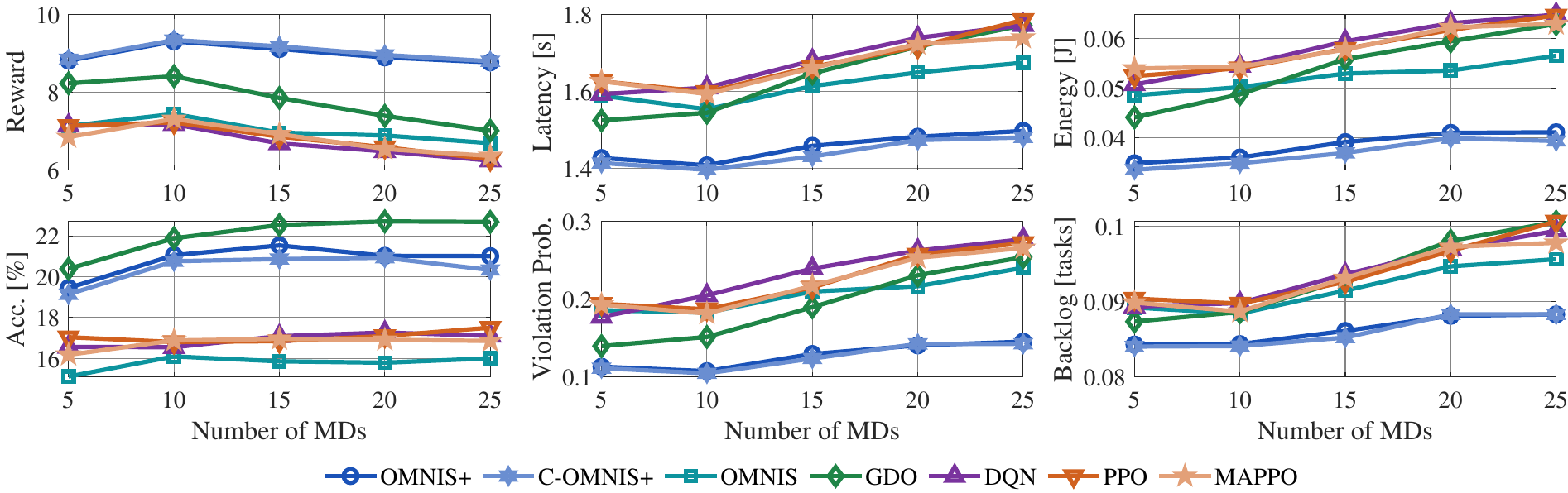}
	\caption{Performance of different schemes under varying numbers of MDs. The top row shows reward, average delay, and energy from left to right. The bottom row shows accuracy, violation probability, and backlog from left to right.}
	\label{fig:diff_user_num}
\end{figure*}

The bottom panel of Fig.\,\ref{fig:convergence} reports per-slot control plane running time on a workstation with dual Intel Xeon Platinum 8470Q CPUs, $750$\,GB RAM, and an NVIDIA GeForce RTX~$5090$ GPU. Each bar stacks four parts. {\em Action selection} is the time to score candidate arms and pick the branch--cell action. {\em Model update} is the time to refresh the accuracy GP or the DRL policy from completed-task feedback. {\em Interaction} is the modeled MD--ES exchange of local delay/energy reports and edge completion reports that all schemes use to keep the shared compute predictors updated. {\em Resource allocation} is the per-cell BCD that assigns bandwidth, GPU frequency, and MCS. The main insight is that distributed control is much cheaper than centralized control. OMNIS+ costs $13.3$\,ms/slot, which is far below the $1$\,s slot length $\delta$, because each MD scores arms and runs the accuracy GP in parallel and only reports its chosen action. C-OMNIS+ uses the same score but is centralized. One controller scores MDs one by one, rescores collisions on the same cell, and each MD uploads a full observation instead of an action. C-OMNIS+ therefore costs $20.6$\,ms/slot, about $1.55{\times}$ that of OMNIS+, for only about a $1\%$ reward gain on the same causal score. Centralized DQN and PPO spend even more time on online model updates. MAPPO is lighter with local actors, but its reward still lags OMNIS+. Therefore, OMNIS+ almost matches C-OMNIS+ in reward while saving much running time, which is suitable for online RAN control.

{\em Average performance under different SNRs.} Fig.\,\ref{fig:diff_snr} reports average performance for target best cell SNR from $0$ to $10$\,dB, averaged over $3$ independent runs. A higher SNR raises the instantaneous SINR of each link between an MD and a cell, which increases the feasible MCS and spectral efficiency and lowers the block error rate. Consequently, uplink transmissions need fewer time slots: delay and energy go down while accuracy rises and QoS violations fall. OMNIS+ achieves the highest reward and the most stable accuracy gains. With better channels and moderate queues, its accuracy GP predicts higher expected accuracy from wider branches, while the online delay, energy, and backlog scores continue to reject overloaded cells. Selections therefore move toward wider models such as Standard-12 and Box-6 and toward stronger Top-$L$ associations. Relative to the conference-style scalar-utility learner OMNIS, OMNIS+ improves inference accuracy by up to $42.5\%$ over this SNR sweep (peak gains near $10$\,dB). OMNIS improves more slowly because its single utility GP mixes accuracy with soft QoS penalties. C-OMNIS+ closely follows OMNIS+ under the same score. GDO selects the arm with the highest predicted accuracy among QoS feasible candidates but has no backlog trade-off. DRL baselines improve with SNR but remain far below OMNIS+.

\begin{figure*}[tbp]
	\centering
	\includegraphics[width=\textwidth]{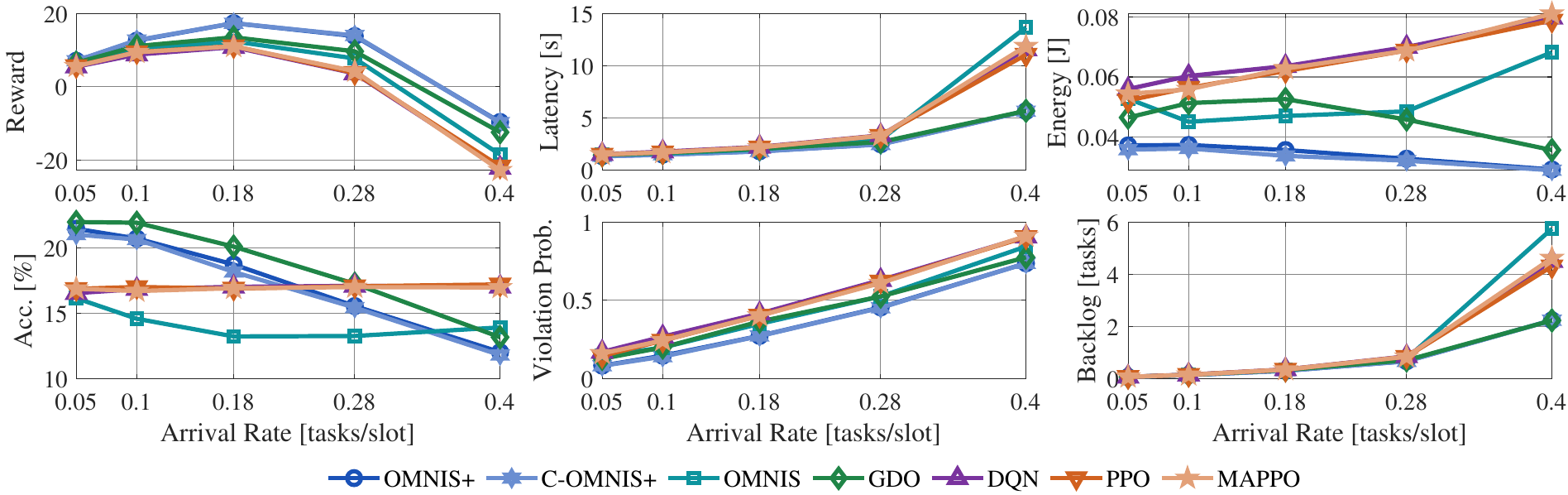}
	\caption{Performance of different schemes under varying task arrival rates. The top row shows reward, average delay, and energy from left to right. The bottom row shows accuracy, violation probability, and backlog from left to right.}
	\label{fig:diff_arrival}
\end{figure*}

\begin{figure*}[thbp]
	\centering
	\includegraphics[width=\textwidth]{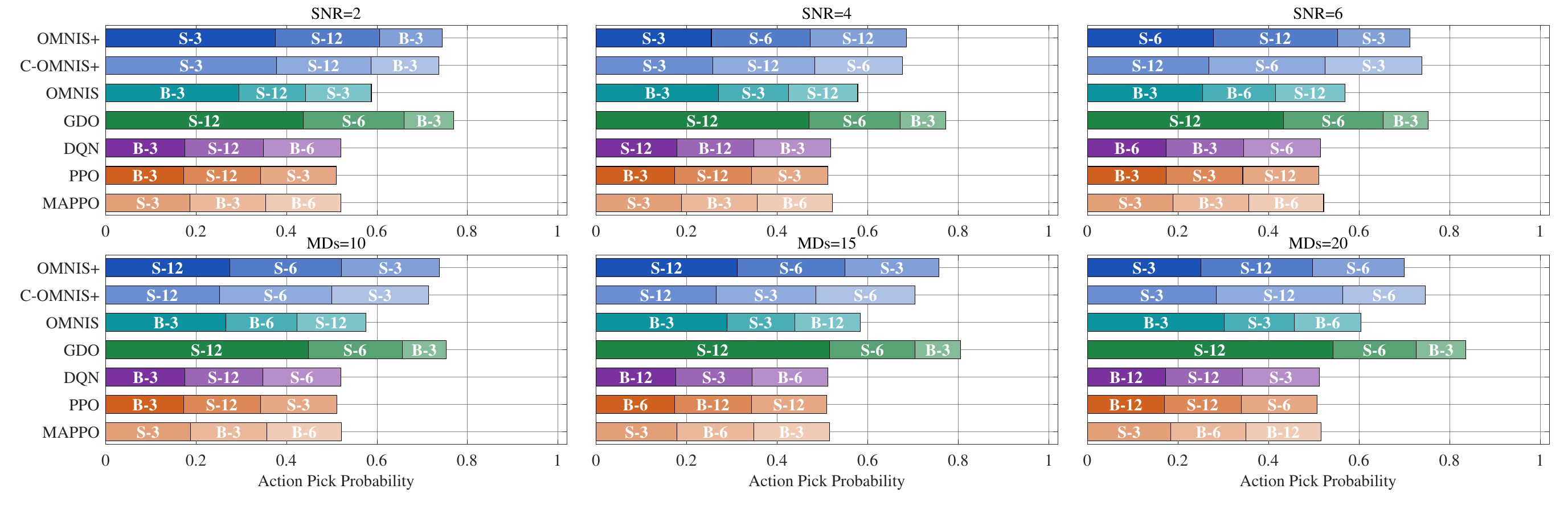}
	\caption{Action pick probabilities of different schemes under varying parameters. The top row corresponds to SNR $=2,4,6$\,dB from left to right. The bottom row corresponds to $M=10,15,20$ from left to right.}
	\label{fig:action_freq}
\end{figure*}

{\em Average performance under different MD numbers.} Fig.\,\ref{fig:diff_user_num} shows system performance across varying numbers of MDs $M$ from $5$ to $25$, averaged over $3$ independent runs. Increasing $M$ raises cell contention and reduces per-MD bandwidth and GPU allocations. At light load, the mean Lyapunov reward can first rise near $M{=}10$ as more MDs finish tasks; beyond that point, delays grow, QoS penalties grow, and reward falls. Uplink energy rises due to longer transmission under contention, while computing energy per task falls as ES GPU resources are shared among more tasks. As delay and energy budgets become harder to meet, violation rates and backlogs rise. Queue-aware controllers then move toward compressed branches: OMNIS+ accuracy first improves with more learning samples, then levels off or slightly decreases once compression dominates. OMNIS+ keeps the best reward and violation trade-off across all $M$ because it scores accuracy separately from online backlog drift. As $M$ grows, each MD can switch to a compressed branch or a less loaded Top-$L$ cell without treating congestion as utility noise. In contrast, OMNIS suffers greater reward loss and higher violations because its scalar GP model treats contention penalties as low utility for wider branches, which leads to overly conservative decisions. C-OMNIS+ follows OMNIS+ closely under the same objective and resolves cell conflicts by local shared-pool rescoring rather than cell reassignment. GDO maximizes predicted accuracy only among predicted QoS-feasible arms and reacts weakly to backlog growth. DRL baselines degrade further as $M$ increases because credit assignment is hard under coupled queues and interference. At $M{=}25$, OMNIS+ reduces the QoS violation probability by $1.75{\times}$ relative to GDO, while keeping a better accuracy--violation trade-off.

{\em Performance under different task arrival rates.} Fig.\,\ref{fig:diff_arrival} evaluates system performance across Poisson arrival rates $\alpha$ (with $\alpha_m{=}\alpha$ for all $m$), averaged over $3$ independent runs. As $\alpha$ increases, tasks arrive faster than they are processed, so the composite backlog $\tilde{Q}_m(t)$ builds up, delay and energy violations rise, and the backlog score $\Delta_m$ weighs more heavily in $Vu+g^{d}\Delta$. Mean reward therefore first rises with more completions at moderate $\alpha$, then falls sharply once backlog and QoS penalties dominate, and turns negative at the highest $\alpha$. Consequently, all schemes move toward smaller payloads, though their mechanisms differ. Under default $V{=}10$, OMNIS+ keeps higher mean rewards than distributed baselines at moderate arrivals. When backlogs rise, its explicit drift score favors compressed branches and safer cells and gives up some accuracy to limit queue growth; this compression also yields an energy curve that first rises and then falls for OMNIS+, C-OMNIS+, and GDO, whereas DRL energy keeps climbing. OMNIS yields lower rewards with higher backlogs because it must estimate this trade-off from noisy scalar samples. C-OMNIS+ inherits this drift-aware score. When admitted tasks compete for the same cell, its capped bandwidth forecast further favors smaller payloads. In contrast, GDO maximizes predicted accuracy subject only to delay and energy feasibility gates: without an explicit backlog term, it has lower peak rewards and higher violations under arrival surges, even though those gates still limit its backlog relative to DRL. Likewise, DRL baselines optimize scalar team returns without a drift term, so their backlogs grow much larger than those of OMNIS+ at high $\alpha$.

{\em Action selection probabilities under varying parameters.} Fig.\,\ref{fig:action_freq} shows the measured branch selection frequencies under different SNRs and MD counts, averaged over $3$ independent runs, which explains the metric trends above. When SNR increases or $M$ decreases, OMNIS+ assigns higher probability to Standard-12 and Box-6. Better spectral efficiency and lighter contention shorten transmission time for a given payload, so the online QoS and drift scores leave room for higher-accuracy branches. When SNR is poor or $M$ is large, OMNIS+ moves mass to compressed models such as Box-3, which reduces airtime, decoding failures, and bit-queue growth at the cost of lower clean accuracy. This switch with channel and load explains why reward and accuracy improve with SNR, and why violation growth stays moderate as $M$ increases. OMNIS shows a similar compression bias but stays more focused on Box-3 at moderate SNR, which matches its slower return to wide branches after early backlog penalties. GDO places more mass on heavier branches whenever they remain QoS feasible, reflecting accuracy maximization without a backlog term. DQN, PPO, and MAPPO spread probability more evenly across branches, which shows weaker focus under delayed scalar rewards. C-OMNIS+ closely matches the OMNIS+ histogram, with only local branch changes when multiple MDs collide on a cell. Overall, these results show that OMNIS+ adapts its branch selection strategy as the radio and load context changes, and keeps a clearer accuracy--QoS trade-off than scalar-utility or greedy baselines.

\begin{figure}[tbp]
	\centering
	\includegraphics[width=\columnwidth]{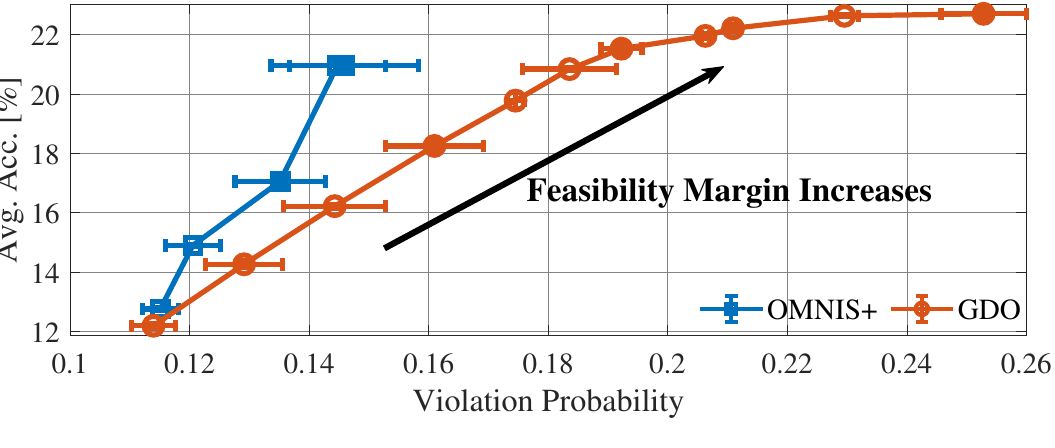}
	\caption{Accuracy and violation trade-off of OMNIS+ and GDO.}
	\label{fig:acc_vio}
\end{figure}

{\em Accuracy and violation trade-off.} 
GDO can reach higher raw accuracy than OMNIS+ under a loose feasibility gate, but its violation probability is also higher. To show that OMNIS+ offers a better accuracy--violation trade-off, Fig.\,\ref{fig:acc_vio} compares OMNIS+ and GDO under fixed shared parameter settings, averaged over $3$ independent runs. We sweep the feasibility margin $\varpi$ on predicted delay and energy, with $\hat{\tau}^{f}{\le}\varpi\tau^{h}_{m,n}$ and $\hat{E}^{f}{\le}\varpi E^{h}_{m,n}$, so that the realized violation ranges overlap. Under a loose gate, GDO reaches higher raw accuracy by freely admitting heavy branches, but the realized violation rate is also much higher. Tightening the gate reduces both accuracy and violation and moves GDO into the same accuracy and violation region as OMNIS+ for comparison. Across the overlapping violation band from about $0.12$ to $0.15$, OMNIS+ achieves higher accuracy under the same violation probability. At matched accuracy, OMNIS+ yields a lower violation probability. Thus, OMNIS+ is preferable when both task quality and QoS reliability matter.

\section{Conclusions}\label{sec:conclusions}
We proposed OMNIS+, a semantic RAN slicing framework for multi-cell edge computing. By combining dynamic split neural networks with adaptive quantization, OMNIS+ provides flexible payload and computing options that remain resilient to bit errors under varying radio conditions. In particular, we proposed box quantization to improve the trade-off between compression rate and transmission robustness for task offloading. To address partial observability and differing objectives of network nodes, we formulated two coupled problems for MDs and ESs. Each MD maximizes inference accuracy under QoS and queue-stability constraints by jointly selecting the split-DNN branch and a serving cell through a causal contextual MAB under a backlog-aware Lyapunov objective. Each ES maximizes a max--min accuracy score under per-cell capacity limits and, online, solves $\mathrm{P2'}$ via BCD over bandwidth, GPU frequency, and MCS. Results show that, compared with state-of-the-art baselines, OMNIS+ reduces the QoS violation probability by up to $1.75{\times}$ while achieving a better accuracy--violation trade-off.

\bibliographystyle{IEEEtran}
\bibliography{ref_short}

@String{Computing = "Computing" }

@String{Computer = "{IEEE} Computer" }

@ARTICLE{Langtian2024,
  author={Qin, Langtian and Lu, Hancheng and Chen, Yuang and Chong, Baolin and Wu, Feng},
  journal={IEEE Transactions on Mobile Computing}, 
  title={Toward Decentralized Task Offloading and Resource Allocation in User-Centric MEC}, 
  year={2024},
  volume={23},
  number={12},
  pages={11807-11823}}

@ARTICLE{Hongyu2020,
  author={Xiang, Hongyu and Yan, Shi and Peng, Mugen},
  journal={IEEE Transactions on Wireless Communications}, 
  title={A Realization of Fog-RAN Slicing via Deep Reinforcement Learning}, 
  year={2020},
  volume={19},
  number={4},
  pages={2515-2527},
  doi={10.1109/TWC.2020.2965927}}

@ARTICLE{Salvatore2018,
  author={D’Oro, Salvatore and Restuccia, Francesco and Melodia, Tommaso and Palazzo, Sergio},
  journal={IEEE/ACM Transactions on Networking}, 
  title={Low-Complexity Distributed Radio Access Network Slicing: Algorithms and Experimental Results}, 
  year={2018},
  volume={26},
  number={6},
  pages={2815-2828},
  doi={10.1109/TNET.2018.2878965}}

@ARTICLE{Yao2021,
  author={Sun, Yao and Qin, Shuang and Feng, Gang and Zhang, Lei and Imran, Muhammad Ali},
  journal={IEEE Transactions on Mobile Computing}, 
  title={Service Provisioning Framework for RAN Slicing: User Admissibility, Slice Association and Bandwidth Allocation}, 
  year={2021},
  volume={20},
  number={12},
  pages={3409-3422},
  doi={10.1109/TMC.2020.3000657}}

@ARTICLE{Salvatore2021,
  author={D’Oro, Salvatore and Bonati, Leonardo and Restuccia, Francesco and Melodia, Tommaso},
  journal={IEEE/ACM Transactions on Networking}, 
  title={Coordinated 5G Network Slicing: How Constructive Interference Can Boost Network Throughput}, 
  year={2021},
  volume={29},
  number={4},
  pages={1881-1894},
  doi={10.1109/TNET.2021.3073272}}

@ARTICLE{Veloso2024,
  author={Nahum, Cleverson Veloso and Lopes, Victor Hugo L. and Dreifuerst, Ryan M. and Batista, Pedro and Correa, Ilan and Cardoso, Kleber Vieira and Klautau, Aldebaro and Heath, Robert W.},
  journal={IEEE Transactions on Wireless Communications}, 
  title={Intent-Aware Radio Resource Scheduling in a RAN Slicing Scenario Using Reinforcement Learning}, 
  year={2024},
  volume={23},
  number={3},
  pages={2253-2267},
  doi={10.1109/TWC.2023.3297014}}

@ARTICLE{Mehdi2022,
  author={Setayesh, Mehdi and Bahrami, Shahab and Wong, Vincent W.S.},
  journal={IEEE Transactions on Wireless Communications}, 
  title={Resource Slicing for eMBB and URLLC Services in Radio Access Network Using Hierarchical Deep Learning}, 
  year={2022},
  volume={21},
  number={11},
  pages={8950-8966},
  doi={10.1109/TWC.2022.3171264}}

@ARTICLE{Kai2022,
  author={Yu, Kai and Zhou, Haibo and Tang, Zhixuan and Shen, Xuemin and Hou, Fen},
  journal={IEEE Transactions on Wireless Communications}, 
  title={Deep Reinforcement Learning-Based RAN Slicing for UL/DL Decoupled Cellular V2X}, 
  year={2022},
  volume={21},
  number={5},
  pages={3523-3535},
  doi={10.1109/TWC.2021.3122941}}

@inproceedings{Kai2022Mobicom,
author = {Huang, Kai and Gao, Wei},
title = {Real-time neural network inference on extremely weak devices: agile offloading with explainable AI},
year = {2022},
publisher = {ACM},
address = {New York, NY, USA},
doi = {10.1145/3495243.3560551},
pages = {200–213},
booktitle= {ACM MobiCom}
}

@ARTICLE{Guangji2023,
  author={Chen, Guangji and Wu, Qingqing and Liu, Ruiqi and Wu, Jingxian and Fang, Chao},
  journal={IEEE Journal on Selected Areas in Communications}, 
  title={IRS Aided MEC Systems With Binary Offloading: A Unified Framework for Dynamic IRS Beamforming}, 
  year={2023},
  volume={41},
  number={2},
  pages={349-365},
  doi={10.1109/JSAC.2022.3228605}}

@ARTICLE{Liang2023,
  author={Zhao, Liang and Zhao, Zijia and Zhang, Enchao and Hawbani, Ammar and Al-Dubai, Ahmed Y. and Tan, Zhiyuan and Hussain, Amir},
  journal={IEEE Journal on Selected Areas in Communications}, 
  title={A Digital Twin-Assisted Intelligent Partial Offloading Approach for Vehicular Edge Computing}, 
  year={2023},
  volume={41},
  number={11},
  pages={3386-3400},
  doi={10.1109/JSAC.2023.3310062}}

@ARTICLE{Ying2024,
  author={Chen, Ying and Li, Kaixin and Wu, Yuan and Huang, Jiwei and Zhao, Lian},
  journal={IEEE Transactions on Mobile Computing}, 
  title={Energy Efficient Task Offloading and Resource Allocation in Air-Ground Integrated MEC Systems: A Distributed Online Approach}, 
  year={2024},
  volume={23},
  number={8},
  pages={8129-8142},
  doi={10.1109/TMC.2023.3346431}}

@ARTICLE{Hongbo2023,
  author={Jiang, Hongbo and Dai, Xingxia and Xiao, Zhu and Iyengar, Arun},
  journal={IEEE Transactions on Mobile Computing}, 
  title={Joint Task Offloading and Resource Allocation for Energy-Constrained Mobile Edge Computing}, 
  year={2023},
  volume={22},
  number={7},
  pages={4000-4015},
  doi={10.1109/TMC.2022.3150432}}

@ARTICLE{Xin2021,
  author={Tang, Xin and Chen, Xu and Zeng, Liekang and Yu, Shuai and Chen, Lin},
  journal={IEEE Internet of Things Journal}, 
  title={Joint Multiuser DNN Partitioning and Computational Resource Allocation for Collaborative Edge Intelligence}, 
  year={2021},
  volume={8},
  number={12},
  pages={9511-9522},
  doi={10.1109/JIOT.2020.3010258}}

@ARTICLE{Corrado2024TMC,
  author={Puligheddu, Corrado and Ashdown, Jonathan and Chiasserini, Carla Fabiana and Restuccia, Francesco},
  journal={IEEE Transactions on Mobile Computing}, 
  title={SEM-O-RAN: Semantic O-RAN Slicing for Mobile Edge Offloading of Computer Vision Tasks}, 
  year={2024},
  volume={23},
  number={7},
  pages={7785-7800},
  doi={10.1109/TMC.2023.3339056}}

@ARTICLE{Qifan2023,
  author={Fu, Qifan and Xie, Huiqiang and Qin, Zhijin and Slabaugh, Gregory and Tao, Xiaoming},
  journal={IEEE Wireless Communications Letters}, 
  title={Vector Quantized Semantic Communication System}, 
  year={2023},
  volume={12},
  number={6},
  pages={982-986},
  doi={10.1109/LWC.2023.3255221}}

@ARTICLE{Danlan2023,
  author={Huang, Danlan and Gao, Feifei and Tao, Xiaoming and Du, Qiyuan and Lu, Jianhua},
  journal={IEEE Journal on Selected Areas in Communications}, 
  title={Toward Semantic Communications: Deep Learning-Based Image Semantic Coding}, 
  year={2023},
  volume={41},
  number={1},
  pages={55-71},
  doi={10.1109/JSAC.2022.3221999}}

@ARTICLE{Jiawei2022,
  author={Shao, Jiawei and Mao, Yuyi and Zhang, Jun},
  journal={IEEE Journal on Selected Areas in Communications}, 
  title={Learning Task-Oriented Communication for Edge Inference: An Information Bottleneck Approach}, 
  year={2022},
  volume={40},
  number={1},
  pages={197-211},
  doi={10.1109/JSAC.2021.3126087}}

@INPROCEEDINGS{Kaiming2016,
  author={He, Kaiming and others},
  booktitle={IEEE Conference on Computer Vision and Pattern Recognition (CVPR)}, 
  title={Deep Residual Learning for Image Recognition}, 
  year={2016},
  pages={770-778},
  doi={10.1109/CVPR.2016.90}}

@INPROCEEDINGS{Corrado2024ICDCS,
  author={Puligheddu, Corrado and Varshney, Nancy and Hassan, Tanzil and Ashdown, Jonathan and Restuccia, Francesco and Chiasserini, Carla Fabiana},
  booktitle={IEEE 44th International Conference on Distributed Computing Systems (ICDCS)}, 
  title={OffloaDNN: Shaping DNNs for Scalable Offloading of Computer Vision Tasks at the Edge}, 
  year={2024},
  pages={624-634},
  doi={10.1109/ICDCS60910.2024.00064}}

@INPROCEEDINGS{Yoshitomo2022,
  author={Matsubara, Yoshitomo and Callegaro, Davide and Singh, Sameer and Levorato, Marco and Restuccia, Francesco},
  booktitle={IEEE 23rd International Symposium on a World of Wireless, Mobile and Multimedia Networks (WoWMoM)}, 
  title={BottleFit: Learning Compressed Representations in Deep Neural Networks for Effective and Efficient Split Computing}, 
  year={2022},
  pages={337-346},
  doi={10.1109/WoWMoM54355.2022.00032}}

@inproceedings{matsubara2022supervised,
  title={{Supervised Compression for Resource-Constrained Edge Computing Systems}},
  author={Matsubara, Yoshitomo and Yang, Ruihan and Levorato, Marco and Mandt, Stephan},
  booktitle={IEEE/CVF Winter Conference on Applications of Computer Vision (WACV)},
  pages={2685-2695},
  month={January},
  year={2022}
}

@inproceedings{Jacob2018,
  title={{SQuantization and Training of Neural Networks for Efficient
Integer-Arithmetic-Only Inference}},
  author={Jacob, Benoit and Skirmantas, Kligys and Chen, Bo and Zhu, Menglong and Tang, Matthew and Howard, Andrew  and Adam, Hartwig and Kalenichenko, Dmitry},
  booktitle={IEEE/CVF Winter Conference on Applications of Computer Vision (WACV)},
  pages={2704-2713},
  year={2018}
}

@INPROCEEDINGS{Ayala2021INFOCOM,
  author={Ayala-Romero, Jose A. and Garcia-Saavedra, Andres and Costa-Perez, Xavier and Iosifidis, George},
  booktitle={IEEE Conference on Computer Communications (INFOCOM)}, 
  title={Bayesian Online Learning for Energy-Aware Resource Orchestration in Virtualized RANs}, 
  year={2021},
  pages={1-10},
  doi={10.1109/INFOCOM42981.2021.9488845}}

@inproceedings{Krause2011,
author = {Krause, Andreas and Ong, Cheng Soon},
title = {Contextual Gaussian process bandit optimization},
year = {2011},
publisher = {Curran Associates Inc.},
address = {Red Hook, NY, USA},
booktitle = {25th International Conference on Neural Information Processing Systems},
series = {NIPS'11}
}

@ARTICLE{Zelin2024,
  author={Ji, Zelin and Qin, Zhijin},
  journal={IEEE Journal of Selected Topics in Signal Processing}, 
  title={Computational Offloading in Semantic-Aware Cloud-Edge-End Collaborative Networks}, 
  year={2024},
  volume={18},
  number={7},
  pages={1235-1248},
  doi={10.1109/JSTSP.2024.3433387}}

@ARTICLE{Wen2021,
  author={Wu, Wen and Chen, Nan and Zhou, Conghao and Li, Mushu and Shen, Xuemin and Zhuang, Weihua and Li, Xu},
  journal={IEEE Journal on Selected Areas in Communications}, 
  title={Dynamic RAN Slicing for Service-Oriented Vehicular Networks via Constrained Learning}, 
  year={2021},
  volume={39},
  number={7},
  pages={2076-2089},
  doi={10.1109/JSAC.2020.3041405}}

@inproceedings{COCO,
  author    = {Lin, Tsung-Yi and Maire, Michael and Belongie, Serge and Hays, James and Perona, Pietro and Ramanan, Deva and Doll{\'a}r, Piotr and Zitnick, C. Lawrence},
  title     = {Microsoft {COCO}: Common Objects in Context},
  booktitle = {European Conference on Computer Vision ({ECCV})},
  year      = {2014},
  pages     = {740--755},
  doi       = {10.1007/978-3-319-10602-1_48}
}

@misc{rANS_entropy,
  author       = {Duda, Jarek},
  title        = {Asymmetric numeral systems: entropy coding combining speed of {Huffman} coding with compression rate of arithmetic coding},
  year         = {2013},
  howpublished = {arXiv:1311.2540}
}

@article{split_computing_survey,
author = {Matsubara, Yoshitomo and Levorato, Marco and Restuccia, Francesco},
title = {Split Computing and Early Exiting for Deep Learning Applications: Survey and Research Challenges},
year = {2022},
publisher = {ACM},
address = {New York, NY, USA},
volume = {55},
number = {5},
doi = {10.1145/3527155},
journal = {ACM Comput. Surv.},
month = dec
}

@techreport{3gpp-ts38214,
  author       = {{3GPP}},
  title        = {{NR; Physical layer procedures for data}},
  institution  = {3rd Generation Partnership Project (3GPP)},
  type         = {Technical Specification (TS)},
  number       = {38.214},
  year         = {2024},
  note         = {V18.4.0, Release 18}
}

@misc{Hoydis2022Sionna,
  author       = {Hoydis, Jakob and Cammerer, Sebastian and A{\"\i}t Aoudia, Fay{\c{c}}al and Vem, Avinash and Binder, Nikolaus and Marcus, Guillermo and Keller, Alexander},
  title        = {{Sionna}: An Open-Source Library for Next-Generation Physical Layer Research},
  year         = {2022},
  howpublished = {arXiv:2203.11854}
}

@ARTICLE{Qin2024,
  author={Qin, Langtian and Lu, Hancheng and Chen, Yuang and Chong, Baolin and Wu, Feng},
  journal={IEEE Transactions on Mobile Computing}, 
  title={Toward Decentralized Task Offloading and Resource Allocation in User-Centric MEC}, 
  year={2024},
  volume={23},
  number={12},
  pages={11807-11823},
  doi={10.1109/TMC.2024.3399766}}

@INPROCEEDINGS{Qin2026,
  author={Qin, Langtian and Harshbarger, Ian Andrew and Nasraoui, Leïla and Chiasserini, Carla Fabiana and Levorato, Marco},
  booktitle={IEEE INFOCOM 2026 - IEEE Conference on Computer Communications}, 
  title={OMNIS: Semantic RAN Slicing via Dynamic Split Neural Networks}, 
  year={2026},
  pages={1-10},
  doi={10.1109/INFOCOM59046.2026.11571606}}

@ARTICLE{Mhatre2025,
  author={Mhatre, Suvidha and Adelantado, Ferran and Ramantas, Kostas and Verikoukis, Christos},
  journal={IEEE Transactions on Vehicular Technology}, 
  title={Intelligent QoS-Aware Slice Resource Allocation With User Association Parameterization for Beyond 5G O-RAN-Based Architecture Using DRL}, 
  year={2025},
  volume={74},
  number={2},
  pages={3096-3109},
  doi={10.1109/TVT.2024.3483288}}

@ARTICLE{Chen2024,
  author={Chen, Xing and Hu, Shengxi and Yu, Chujia and Chen, Zheyi and Min, Geyong},
  journal={IEEE Transactions on Parallel and Distributed Systems}, 
  title={Real-Time Offloading for Dependent and Parallel Tasks in Cloud-Edge Environments Using Deep Reinforcement Learning}, 
  year={2024},
  volume={35},
  number={3},
  pages={391-404},
  doi={10.1109/TPDS.2023.3349177}}

@ARTICLE{Nguyen2024,
  author={Nguyen, Van-Dinh and Vu, Thang X. and Nguyen, Nhan Thanh and Nguyen, Dinh C. and Juntti, Markku and Luong, Nguyen Cong and Hoang, Dinh Thai and Nguyen, Diep N. and Chatzinotas, Symeon},
  journal={IEEE Journal on Selected Areas in Communications}, 
  title={Network-Aided Intelligent Traffic Steering in 6G O-RAN: A Multi-Layer Optimization Framework}, 
  year={2024},
  volume={42},
  number={2},
  pages={389-405},
  doi={10.1109/JSAC.2023.3336183}}

@ARTICLE{Xie2025,
  author={Xie, Bo and Cui, Haixia and He, Yejun and Guizani, Mohsen},
  journal={IEEE Transactions on Cognitive Communications and Networking}, 
  title={A Deep Reinforcement Learning Approach for Dependent Task Offloading in Multi-Access Edge Computing}, 
  year={2025},
  volume={11},
  number={4},
  pages={2601-2617},
  doi={10.1109/TCCN.2024.3520961}}

@ARTICLE{Li2025,
  author={Li, Han and Xiong, Ke and Lu, Yuping and Chen, Wei and Fan, Pingyi and Letaief, Khaled Ben},
  journal={IEEE Transactions on Mobile Computing}, 
  title={Collaborative Task Offloading and Resource Allocation in Small-Cell MEC: A Multi-Agent PPO-Based Scheme}, 
  year={2025},
  volume={24},
  number={3},
  pages={2346-2359},
  doi={10.1109/TMC.2024.3496536}}

\end{document}